\def\year{2022}\relax
\documentclass[letterpaper]{article} 
\usepackage{aaai22}  
\usepackage{times}  
\usepackage{helvet}  
\usepackage{courier}  
\usepackage[hyphens]{url}  
\usepackage{graphicx} 
\usepackage{natbib}  
\usepackage{caption} 
\DeclareCaptionStyle{ruled}{labelfont=normalfont,labelsep=colon,strut=off} 
\usepackage{algorithm}
\usepackage{algorithmic}

\usepackage{newfloat}
\usepackage{listings}
\floatstyle{ruled}
\newfloat{listing}{tb}{lst}{}
\floatname{listing}{Listing}
\usepackage{subfigure, amsmath, bbm, amsfonts, multirow, color}

\newcommand{\nn}{\nonumber}
\newcommand{\secref}[1]{Section~\ref{#1}}
\newcommand{\figref}[1]{Figure~\ref{#1}}
\newcommand{\figrefs}[2]{Figures~\ref{#1} and ~\ref{#2}}
\newcommand{\tabref}[1]{Table~\ref{#1}}

\DeclareRobustCommand\onedot{\futurelet\@let@token\@onedot}
\def\@onedot{\ifx\@let@token.\else.\null\fi\xspace}

\def\bx{\mathbf x}

\def\dsum{\displaystyle\sum}

\def\tbf#1{{\textbf{#1}}}
\def\tx#1{{\text{#1}}}

\def\blue#1{{\color{blue}#1}}

\graphicspath{{./figures/}}

\title{Auxiliary Class Based Multiple Choice Learning}
\author {
    Sihwan Kim, 
    Dae Yon Jung, 
    Taejang Park\thanks{Work was done while the author was at Hana TI}
}
\affiliations {
    Hana Institute of Technology, Hana TI \\
    217, Teheran-ro, Gangnam-gu, 
    Seoul, Republic of Korea \\
    \{shnasc.kim, dyjung16, djang000\}@hanafn.com
}

\begin{document}

\maketitle

\begin{abstract}
The merit of ensemble learning lies in having different outputs from many individual models on a single input, $i.e.$, the diversity of the base models. 
The high quality of diversity can be achieved when each model is specialized to different subsets of the whole dataset. 
Moreover, when each model explicitly knows to which subsets it is specialized, the diversity can be maximized. 
In this paper, we propose an advanced ensemble method, called Auxiliary class based Multiple Choice Learning (AMCL), to extremely specialize each model under the framework of multiple choice learning (MCL).
The advancement of AMCL is originated from three novel techniques which control the framework from different directions: 1) the concept of \textbf{auxiliary class} to provide more distinct information about the model's specialization through the labels, 2) the strategy, named \textbf{memory-based assignment}, to determine the association between the inputs and the models, and 3) the \textbf{feature fusion module} to achieve generalized features. 
To demonstrate the performance of our method compared to all variants of MCL methods, we conduct extensive experiments on the image classification and segmentation tasks. 
Overall, the performance of AMCL exceeds all others in most of the public datasets trained with various networks as members of the ensembles. 
\end{abstract}

\section{Introduction}\label{sec:introduction}
Ensemble learning is the process to train multiple models and combine into a single result.
The spotlights to ensemble methods, even until recently from numerous applications in recognition~\cite{app1, app4}, detection~\cite{app2}, and segmentation~\cite{app3}, are derived from their superior predictive performances compared to that of a single model.
This high performance of the ensemble results primarily from the diversity of the multiple models.
If all individual models provide similar outputs, any strategic combination of them cannot reduce the error.
Therefore, achieving high diversity while training an ensemble model,  $i.e.,$ producing a set of various possible outputs, is where the quality of ensemble originates.

Diversity can usually be achieved by alternating either datasets or model configurations, such as hyperparameters and/or initializations, for training each individual model.
Relying on randomization, many of these ensembles require a good number of individual models to fulfill the law of large numbers.
However, rather than differentiating models by randomness,
multiple choice learning (MCL)~\cite{mcl} obtains high diversity by making each model specialize to a certain subset of the dataset.
These specialized models from MCL are directly trained by minimizing a new concept of loss function called \textit{oracle loss}, which assigns each example to one or more, depending on the setting of the constraints, models with the highest accuracy.

Inheriting the concept of the oracle loss, \citet{smcl} applied MCL to ensembles of deep neural networks with a trainable algorithm and named stochastic multiple choice learning (sMCL).
Since each specialized model takes the loss only from a certain set of examples, overconfident predictions to unassigned examples occur by nature. 
To address this overconfidence issue, a series of progressive methods are introduced.
\citet{cmcl} added a regularization term to the oracle loss in order to coerce the predictive probability of an unassigned example to be a uniform distribution.
\citet{vmcl} modified the regularization term to set a fixed margin between the aggregated probability of the ground truth and the others.
In addition, a choice network is appended to the architecture to combine the predictions from all specialized models with different weights.
Both methods substantially improved top-1 error rates while maintaining as good or better oracle errors compared to those of the sMCL method.

Although the accuracies of the ensembles are boosted, all previous methods, utilizing the concept of the oracle loss, do not provide explicit information about the unassigned examples to each specialized model. 
Setting the distribution~\cite{cmcl} or the value range~\cite{vmcl} of the predictive probabilities of unassigned examples is not a direct technique to encourage model specialization. 
The best way to lead the model specialization is to set labels of the unassigned examples with low entropies by directly indicating that the examples are not from any subsets of the model's specialties. 
This clear indication leads every model in the ensemble to an extreme specialization, resulting in highly diversified outputs with lower prediction errors even when simply combined using an averaging scheme.

Inspired by this extreme model specialization, we propose an advanced method, called Auxiliary class based Multiple Choice Learning (AMCL), that expressly aims to train each base model of an ensemble to become the best specialist. 
As the main contributions of AMCL, we firstly develop the \textit{auxiliary class}, an extra element appended to the ground truth label. This class indicates whether or not the input example belongs to a part of the model's specialties. 
From a perspective of training, the auxiliary class helps to 
exclude the examples from the non-specialized classes. 
Secondly, unlike all previous MCL methods that designate each input to a particular model based on the loss value,  $i.e.,$ \textit{loss-based assignment}, we suggest a principled technique, $i.e.,$ \textit{memory-based assignment}, based on the connections between the input examples and the models in the early iterations.
Once the assignment of the inputs to the models are fixed after the first few iterations, the input examples are directly forwarded in the later iterations that helps each model to solely focus on its specialization. 
Finally, we propose a module to fuse low-level features from each model and train the most informative feature through the attention mechanism. This module, named \textit{feature fusion module}, supports to overcome weak generalization which typically occurs under the framework of MCL~\cite{mcl-kd}.

We evaluate AMCL on two representative computer vision tasks: image classification and segmentation. With image classification, extensive experiments are conducted with three large-scale neural networks, VGGNet~\cite{vgg}, ResNet~\cite{resnet}, and GoogLeNet~\cite{googlenet}, on three public image datasets, including CIFAR-10/100~\cite{cifar} and TinyImageNet~\cite{tiny-imagenet}. Compared to the existing MCL methods, AMCL significantly outperforms in both the oracle and the top-1 errors.
We also provide general comparisons with the recent ensemble models, such as the method by~\citet{ennm} (abbreviated as EMADNN), ONE~\cite{one}, and KDCL~\cite{kdcl}, in which AMCL performs better than any of those baselines.
Additionally, the performance on the foreground-background segmentation is evaluated with the iCoseg~\cite{icoseg} dataset.
The overall results show that our proposed method achieves impressive performance in most of the experiments, surpassing the top-1 accuracy as well as the oracle accuracy compared to the other MCL extended methods.

\section{Preliminaries}
\begin{figure*}[!ht]
	\centering
	\includegraphics[width=1.90\columnwidth, height=0.91\columnwidth]{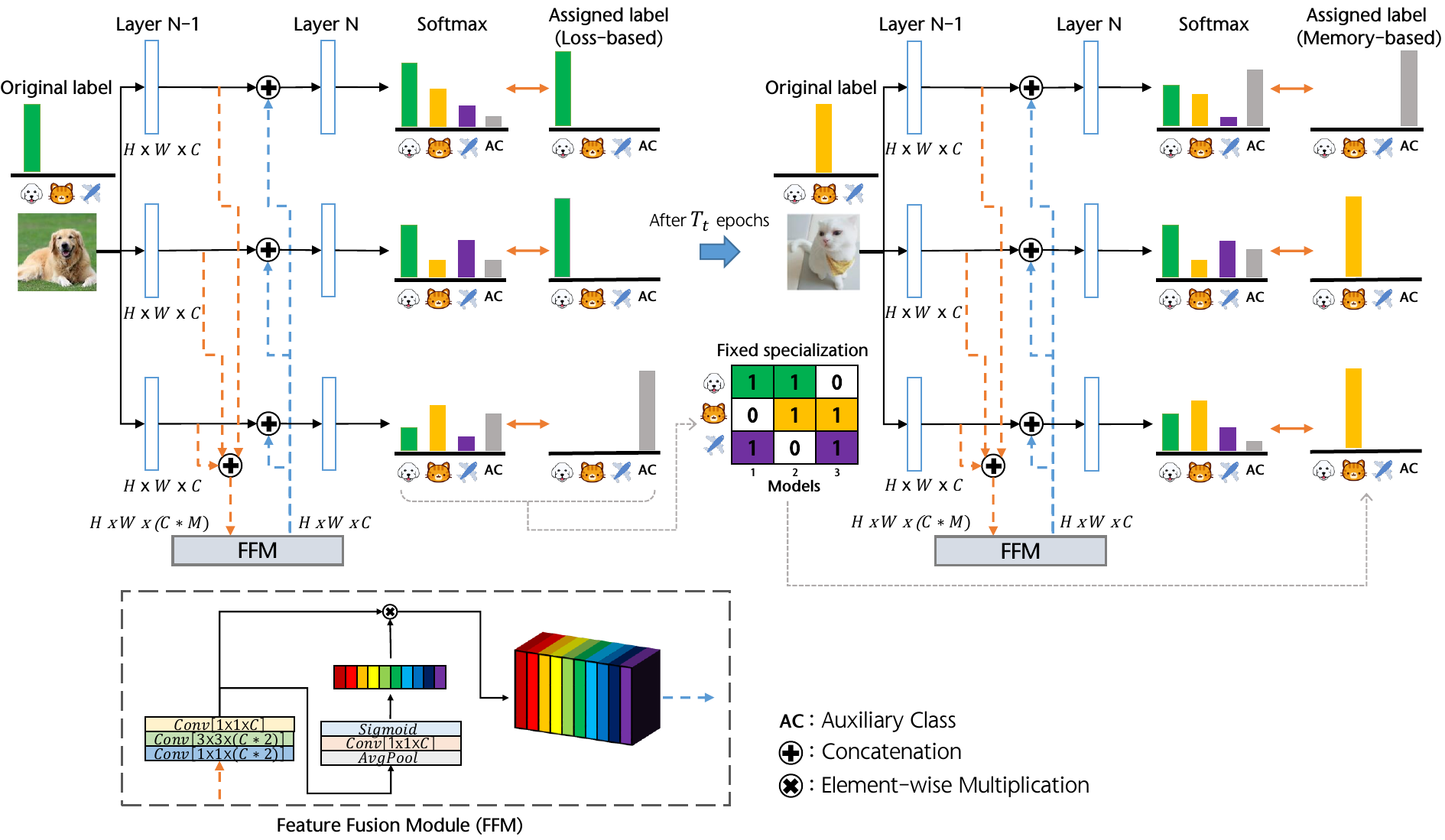}
	\caption{The learning framework of AMCL method with 3 networks specialized to 2 networks per class, $i.e., M = 3$ and $K=2$. All three techniques (auxiliary class, memory-based assignment and feature fusion module) are visualized in this diagram. ``Fixed specialization'' is a matrix indicating the specialization of each model to the classes, which is used in the process of memory-based assignment for training after $T_{\tau}$ epochs.}
	\label{fig:network}
\end{figure*}
In this section, we describe the formulation of the oracle loss from multiple choice learning~\cite{mcl}, on which all MCL related algorithms are based. 
Let $D = \{(\bx_j, y_j)_{j=1}^N\}$ be a dataset, where $\bx_j$ represents a training example, and $y_j$ is the ground truth label, consisting of $N$ i.i.d. observations. Each individual model is denoted by $f_m(\bx),~m=1,\cdots,M$, where $M$ implies the ensemble size.
Formally, a traditional independent ensemble (IE) method trains each model with the following objective:
\begin{equation}\label{eq:ie_loss}
    \footnotesize
	\mathcal{L}_{IE}(D) = \sum_{j=1}^{N}\sum_{m=1}^M \ell\Big(y_j, f_m(\bx_j)\Big),
\end{equation}
where $\ell(\cdot, \cdot)$ implies the task-specific loss function. 
Differently from the traditional IE method, the objective of MCL is to minimize the oracle loss, which accounts only the lowest loss from the predictions of the $M$ models, defined by:
\begin{equation}\label{eq:oracle_loss}
    \footnotesize
	\mathcal{L}_{oracle}(D) = \sum_{j=1}^{N}\min_{m \in \{1,\cdots,M \} } \ell\Big(y_j, f_m(\bx_j)\Big).
\end{equation}
Owing to this loss function, the most accurate model for an individual example $\bx_j$ is determined, and thus, each model in the ensemble is trained to become a specialist to certain subsets or classes of the dataset.

Since the oracle loss~\eqref{eq:oracle_loss} is a non-continuous function, an iterative block coordinate decent algorithm is used to optimize the function.
However, directly applying MCL to an ensemble of deep neural networks is not efficient due to its training costs.
As a solution, Stochastic Multiple Choice Learning (sMCL)~\cite{smcl} proposed a training scheme based on the stochastic gradient descent algorithm to concurrently train ensembles of deep neural networks.
Specifically, sMCL relaxed the objective function to the following integer programming problem:
{\footnotesize
\begin{flalign}\label{eq:smcl_loss}
&\mathcal{L}_{sMCL}(D) = \min_{v_j^m} \sum_{j=1}^{N}\sum_{m=1}^M  v_j^m\ell\Big( y_j, \mathbf{P}_m(\bx_j) \Big), \\
&\qquad\qquad\quad \text{subject to} \quad \sum_{m=1}^M v_j^m = K, \nn
\end{flalign}
}
where $\mathbf{P}_m(\bx_j)$ is the prediction of $m$-th network, and $v_j^m \in \{0,1\}$ is an indicator variable representing the assignment of the input $\bx_j$ to the $m$-th model. 
$K \in \{1,\cdots,M\}$ is the number of specialized models per each input example. 
We note that if $K=1$, the algorithm chooses the single most accurate network for every input and back-propagates only in the corresponding network. 
Consequently, each network is trained and specialized to a certain subset of the dataset. 
When $K=M$, sMCL becomes identical to the IE method.

\section{Auxiliary class based Multiple Choice Learning}
In this section, we describe the technical details and novelties of our proposed training method, consisting of three major parts: (a) auxiliary class, (b) memory-based assignment, and (c) feature fusion module. 
These techniques combined enhance the specialization of each model in the ensembles which, in turn, ultimately promotes the performance in both oracle and top-1 error rates.
We remark that the three newly introduced parts have no restriction on the architecture of each individual models, which lead our method applicable to general fields. 
The overall learning framework of AMCL is illustrated in \figref{fig:network}.

\subsection{Auxiliary Class}\label{sec:aux_class}
The oracle loss~\eqref{eq:oracle_loss} used for MCL is useful for producing diverse solutions in high quality because only the most accurate network corresponding to each input example is trained, and thus, each model eventually focuses on a certain subset of the dataset.
However, this property yields to the overconfidence issue. To address this issue, confident multiple choice learning (CMCL)~\cite{cmcl} proposed a new loss function named $\mathit{confident~oracle~loss}$. 
The confident oracle loss adds a regularization term to the original oracle loss to minimize Kullback-Leibler divergence between the predictive distribution of an unassigned example and a uniform distribution. 

To understand better about the effect of this regularization term, assume that we have an image dataset with two classes: dog and airplane. 
If model `A' and `B' are specialized in each class, dog and airplane, respectively, the predictive probabilities from model `A' for any airplane images will be uniform, $[1/2, 1/2]^T$. 
Thus, CMCL forces each model to be less confident about non-specialized classes, and in terms of performance, it notably reduces the top-1 error while keeping the oracle error similar to that of MCL. 
Nevertheless, the interpretation of the probability is ambiguous since model `A' returns a dog class with the confidence level of 50\% to an airplane image. 

From a human perspective, it makes more sense when model `A' rejects an image of airplane with the confidence level of 100\% because model `A' is specialized in dog, not airplane.
To this end, we add an auxiliary class to efficiently handle the non-specialized classes of each model. 
When model `A' is fully trained along with this auxiliary class, the predictive probability of model `A' for an input of airplane will be $[0, 0, 1]^T$, where the last component is the auxiliary class.
This probability explicitly indicates that the input is definitely not a dog image.
At the inference stage, model `A' regards the probabilities of airplane images as $[0, 0]^T$ through a simple post-process, directly removing the auxiliary class. 
Since model `B' is specialized in the airplane class, the predictive probability to an example of airplane will be $[0, 1]^T$ after the post-processing, and thus, the ensemble prediction via averaging scheme becomes $[0, 1/2]^T$. 
When the prediction is normalized to make the sum equal to 1, AMCL results in $[0, 1]^T$ while CMCL's prediction is $[1/4, 3/4]^T$ by calculating the average between $[1/2, 1/2]^T$ from model `A' and $[0, 1]^T$ from model `B'.
Through this simple yet effective technique, AMCL produces a more intuitive and plausible solution compared to CMCL.

\begin{figure}[!t]
	\centering
	\subfigure [The histogram of dog] {\includegraphics[width=0.47\linewidth, height=0.3\linewidth]{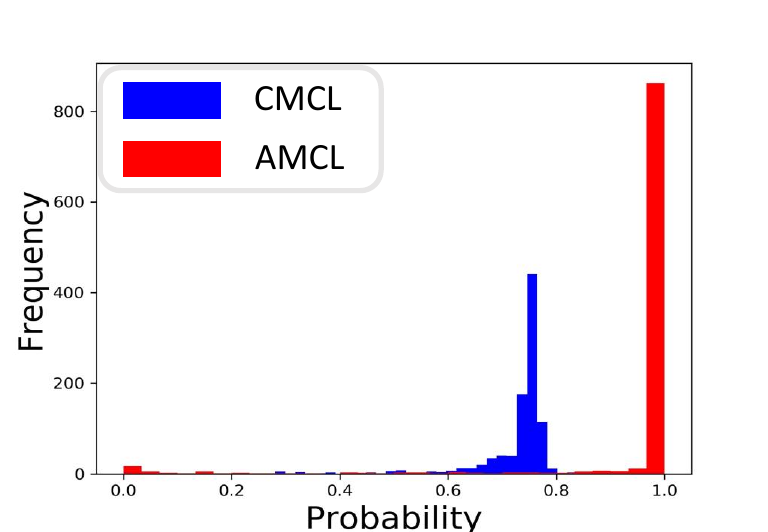}} \quad
	\subfigure [The histogram of airplane ] {\includegraphics[width=0.47\linewidth, height=0.3\linewidth]{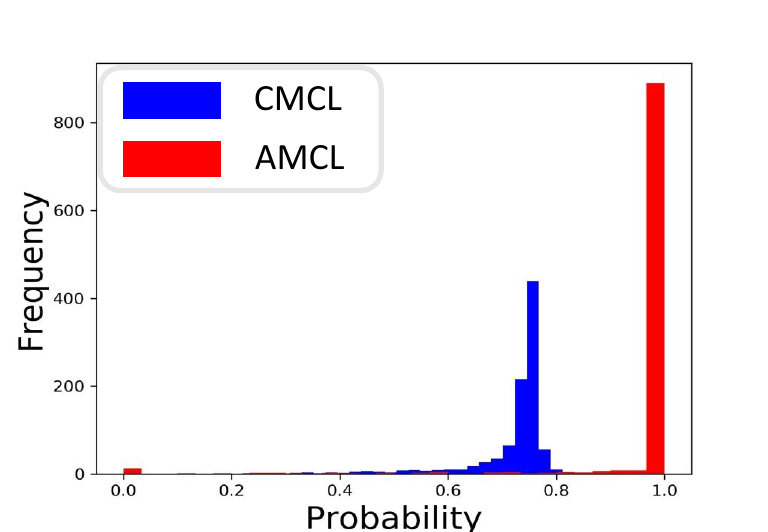}} \quad
	\caption{The histogram of the confidence level in the class of dog and airplane. Each histogram from AMCL and CMCL is in red and blue, respectively. The test dataset consists of 1,000 samples from each class.}
	\label{fig:binary}
\end{figure}

To demonstrate the example mentioned above, we trained an ensemble model with a simple CNN using only two classes, dog and airplane, from CIFAR-10 and compared the confidence level between CMCL and AMCL.
The simple CNN consists of three convolutional layers with 32, 64 and 128 filters followed by a fully-connected layer.
As shown in \figref{fig:binary}, the results indeed show that the confidence levels of both classes in CMCL lie on near 0.75 while those of AMCL lie on near 1.
Additionally, AMCL shows the oracle / top-1 error rates of 0.03$\%$ / 3.12$\%$, which is superior to those of CMCL, 0.21$\%$ / 3.21$\%$.
AMCL achieves up to 85.71$\%$ relative reduction in the oracle error rate from the corresponding model trained with the CMCL scheme.

\subsection{Memory-based Assignment}
Following the constraints of the oracle loss~\eqref{eq:oracle_loss}, only the networks with the lowest loss for the corresponding input examples are trained with the ground truth labels.
Since all the variants of MCL adopt the same oracle loss, this procedure occurs at each training step.
With this strategy, \textit{loss-based assignment}, the examples which do not belong to the specialized class of a model might yield to the lowest loss, and as a consequence, these wrongly assigned examples diminish the purity of the model specialization.

To tackle this problem, we introduce a simple technique to raise the purity of input to model assignment, named \textit{memory-based assignment}. 
The memory-based assignment counts, by online manner, the number of examples per class that belongs to each model within the first few epochs of the training. 
The numbers of examples in some classes becomes incrementally larger than those of the other classes, so after couple epochs, the specialized classes for each model can be naturally decided.
After a set number of epochs, the specialized classes are hard-coded according to the counts of examples that results in the ``Fixed specialization'' matrix in~\figref{fig:network}. 
This strategy helps the model to focus more on a certain subset of the dataset for the rest of the training without any inputs that interrupt the specialization.

To describe in more detail with~\figref{fig:network}, whether to assign the label to the auxiliary class or not is initially decided based on the loss values. 
After the set number of epochs, the specialization of each model is fixed according to the history of the past assignments, which results in the ``Fixed specialization''. 
Therefore, as shown in the right side of the figure, even when the probability of cat turns out to be the lowest given one of the cat images, the assignment of the label to the second model is still cat, and this individual model will be trained to specialize in the images of cats. 

Originated from the objective function of sMCL~\eqref{eq:smcl_loss}, a modified loss for our algorithm implements both of the auxiliary class and the memory-based assignment. 
Let $P_{\theta_m}(\widetilde{y}~|~\bx)$ be the predictive distribution where $\widetilde{y}$ is the output with the auxiliary class appended, and $\theta_m$ denotes the $m$-th model parameters. 
$\mathcal{A}(\widetilde{y})$ is the one-hot encoded vector with only the auxiliary class being 1. 
Also, $t_s$ is the current epoch number, and $T_{\tau}$ denotes the threshold of epochs from loss-based to memory-based.
Then, the objective of AMCL is defined as follows:
{\footnotesize
\begin{flalign}\label{eq:amcl_loss}
	\mathcal{L}_{AMCL}(D)
	=\begin{cases}
		\mathcal{L}_{LBA}(D) & \mbox{if}~t_s \leq T_{\tau}, \\ 
		\mathcal{L}_{MBA}(D) & \mbox{otherwise}, 
	\end{cases} 
\end{flalign}
}
where ${L}_{LBA}(D)$ and ${L}_{MBA}(D)$ are the loss functions from loss-based assignment (LBA) and memory-based assignment (MBA), which are presented below.
{\footnotesize
\begin{flalign}\label{eq:amcl_loss1}
	\begin{aligned}
		&\mathcal{L}_{LBA}(D) =\min_{v_j^m} \dsum_{j=1}^{N}\dsum_{m=1}^M  v_j^m\ell\Big( \widetilde{y}_j, P_{\theta_m}(\widetilde{y}~|~\bx_j) \Big) \\
		&\qquad +\beta (1-v_j^m)D_{KL}\Big( \mathcal{A}(\widetilde{y})~||~P_{\theta_m}(\widetilde{y}~|~\bx_j) \Big) \\ \smallskip 
		&\qquad\qquad\quad \text{subject to} \quad\enspace \dsum_{m=1}^M v_j^m = K,
	\end{aligned}
\end{flalign}
}
and
{\footnotesize
\begin{flalign}\label{eq:amcl_loss2}
	&\mathcal{L}_{MBA}(D) = \dsum_{j=1}^{N}\dsum_{m=1}^M  w_{c(j)}^m\ell\Big( \widetilde{y}_j, P_{\theta_m}(\widetilde{y}~|~\bx_j) \Big) \nn \\
	&\qquad  +\gamma (1-w_{c(j)}^m)D_{KL}\Big( \mathcal{A}(\widetilde{y})~||~P_{\theta_m}(\widetilde{y}~|~\bx_j) \Big),
\end{flalign}
}
where $D_{KL}$ denotes the KL divergence, $\beta$ and $\gamma$ are the penalty parameters, and $v_j^m$ is the indicator variable as defined in~\eqref{eq:smcl_loss}.
$w_{c(j)}^m$ is an element of binary flag matrix of the dimension $(N_c \times M)$ where $c(j)$ denotes the class index for each input example $\bx_j$, and $N_c$ is the total number of classes. 
This matrix is generated by the cumulative sum of $v_j^m$ until the epoch reaches the threshold, $T_{\tau}$.
Each element of the matrix is converted to a binary flag by choosing the top-$K$ counts, row-wise, where $K$ means the set number of models for each class.
Therefore, $w_{c(j)}^m$ implies that the $m$-th model is specialized in the class of the input example $\bx_j$, and this specific class will be assigned to the $m$-th model until the end of the training.
Therefore, AMCL trains each model to become a specialist for certain classes with the purity of high intensity compared to the other MCL methods.


\subsection{Feature Fusion Module}
Only with the auxiliary class and the memory-based assignment, it is challenging to overcome the weak generalization problem from each model under the MCL framework as mentioned by~\citet{mcl-kd}.
When the model specialization is completed with only the auxiliary class and the memory-based assignment, each model is trained without the general information about the whole dataset because all the examples belonging to the non-specialized classes are labeled to a single auxiliary class.
Therefore, to further address this generalization issue, we suggest an additional module, called \textit{feature fusion module}, in the lower level of the network with the architecture similar to the attention networks~\cite{bam, cbam}.
As each model in the ensemble is forced to learn low-level features differently, these features are concatenated and fed into the feature fusion module to generate a more comprehensive but concise feature, which is uniformly provided back to each model for further training.
The fusion module is deployed just after a few layers of each model since the first couple layers tend to learn more common features~\cite{trans_feature}.
Owing to our feature fusion module, the different perspectives about a single input from each ensemble members are combined in the early layer, and all model can learn their own specialized expertise with more generalized features.  

In the previous methods, similar techniques to enhance the low-level features are applied as well.
CMCL proposes the feature sharing scheme, which stochastically shuffles the features among the ensemble models. 
Meanwhile, versatile multiple choice learning (vMCL)~\cite{vmcl} suggests to share the weights of the couple foremost convolutional layers among the models. 
However, in contrast to these previous techniques, feature fusion module approaches this problem in terms of learning the optimal knowledge from all models.
The effect of this module compared to the other techniques is experimentally examined in~\secref{sec:img_class}.

\section{Experiments}\label{sec:exp}
We evaluate the experiments on classification and foreground-background segmentation tasks to verify the effectiveness of our algorithm, AMCL.
The three public image datasets, CIFAR-10/100~\cite{cifar} and TinyImageNet~\cite{tiny-imagenet}, are used for classification, and the iCoseg~\cite{icoseg} dataset is used for segmentation. 
In all experiments, we compare AMCL with others under the same experimental setups for fair comparisons.
In addition to this section, the details about the experimental setups and the hyperparameter studies are provided in the supplementary material.

\subsection{Image Classification}\label{sec:img_class}
We compare all methods using three conventional large CNN models. 
ResNet-18 is used for the ablation and the effect studies while VGGNet-16, ResNet-34, and GoogLeNet are applied for the comprehensive experiments. 
All experiments in this section uses ensembles of 5 networks, $M=5$. 
The experiment with a larger ensemble size is presented in the supplementary material.

Both oracle and top-1 error rates are measured for evaluating the classification performances.
The oracle error is the ratio when all models failed to classify for a given test image, and the top-1 error is calculated by averaging the output probabilities of all models.
We note that a lower oracle error rate implies high diversity but not always the highest top-1 accuracy.

\subsubsection{Ablation study}
We study the contributions of our three major techniques under the comparison with IE and the variants of MCL.
All methods in comparison are trained on CIFAR-10 with an ensemble of 5 ResNet-18. 
As shown in~\tabref{table:ablation}, the performance improves gradually as each technique is incrementally applied.
Through this study, we note that memory-based assignment contributes more to lowering the oracle error while feature fusion module reduces the top-1 error.
When all techniques are applied, AMCL achieves the best performance compared to all others.
We find that AMCL provides 36.3$\%$ and 10.1$\%$ relative error reductions in the oracle and top-1 error rates, respectively, from CMCL, the second best method in terms of harmonic mean of the oracle and top-1 error.

\subsubsection{Analysis of auxiliary class}
\begin{figure*}[t!]
	\centering
	\subfigure [Probability of auxiliary class \label{fig:ac_anal}] {\includegraphics[width=0.25\linewidth, height=0.185\linewidth]{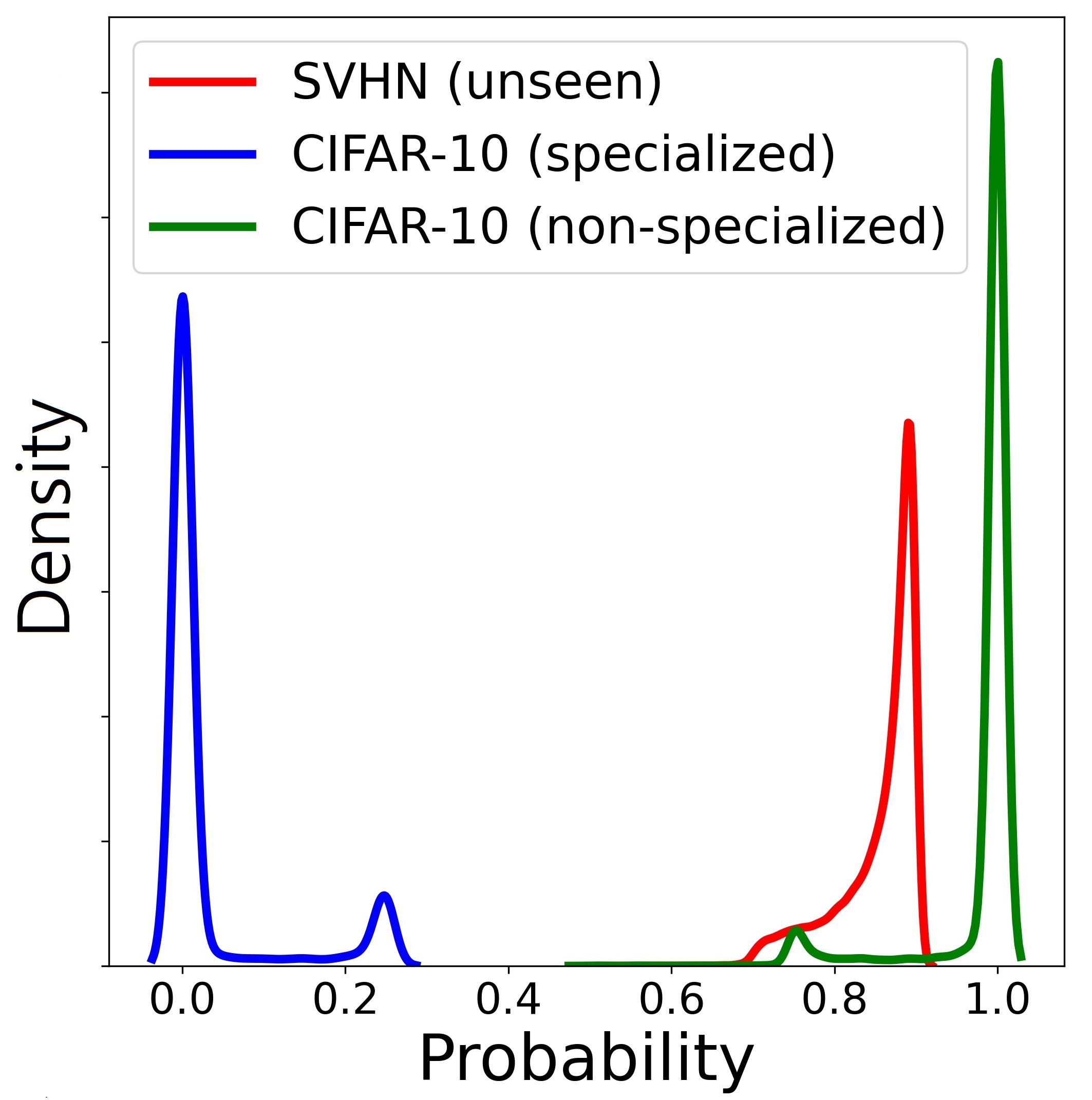}} 
	\subfigure [From specialized classes \label{fig:special_anal}] {\includegraphics[width=0.25\linewidth, height=0.185\linewidth]{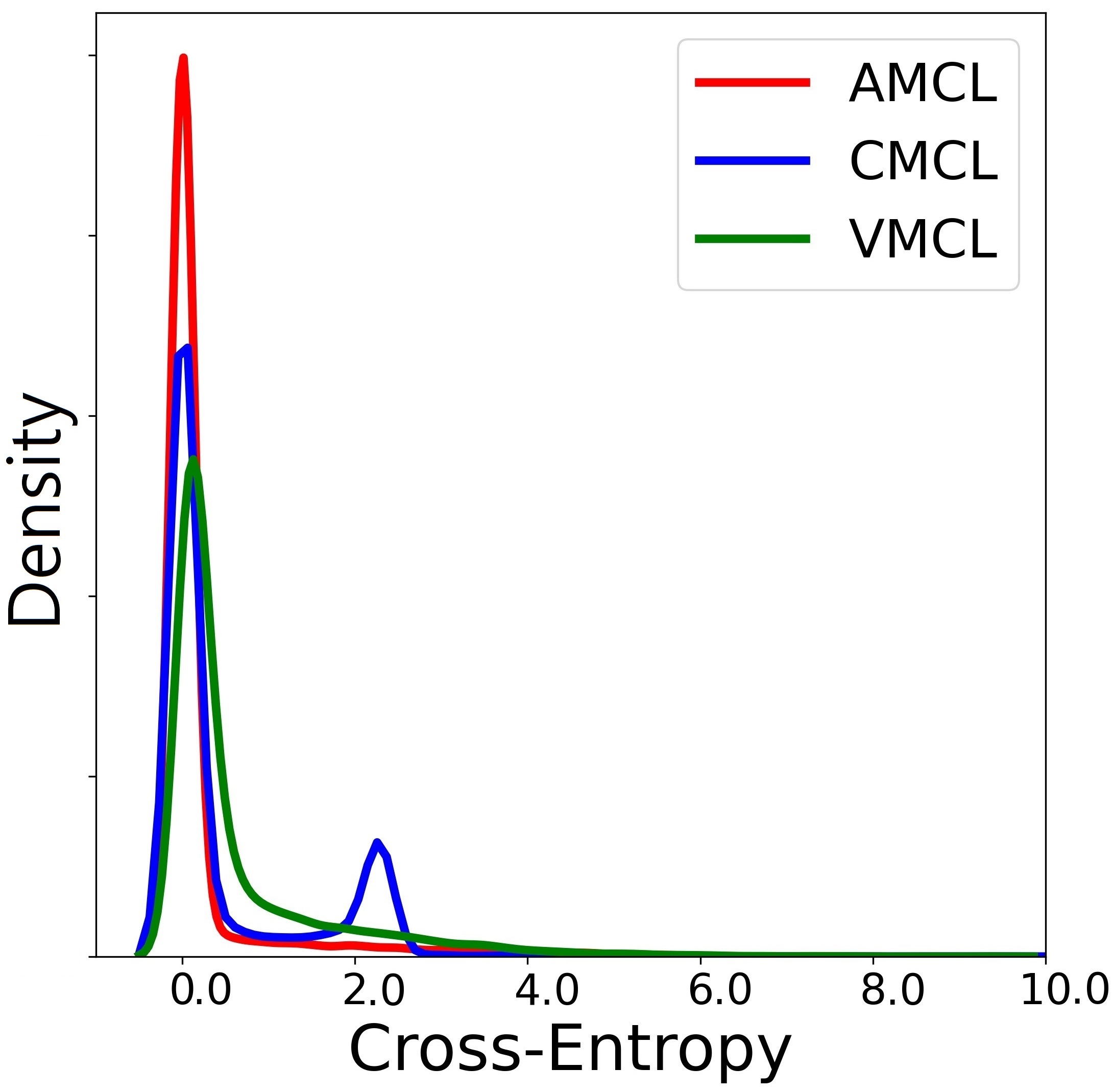}} 
	\subfigure [From non-specialized classes \label{fig:nonspecial_anal}] {\includegraphics[width=0.25\linewidth, height=0.185\linewidth]{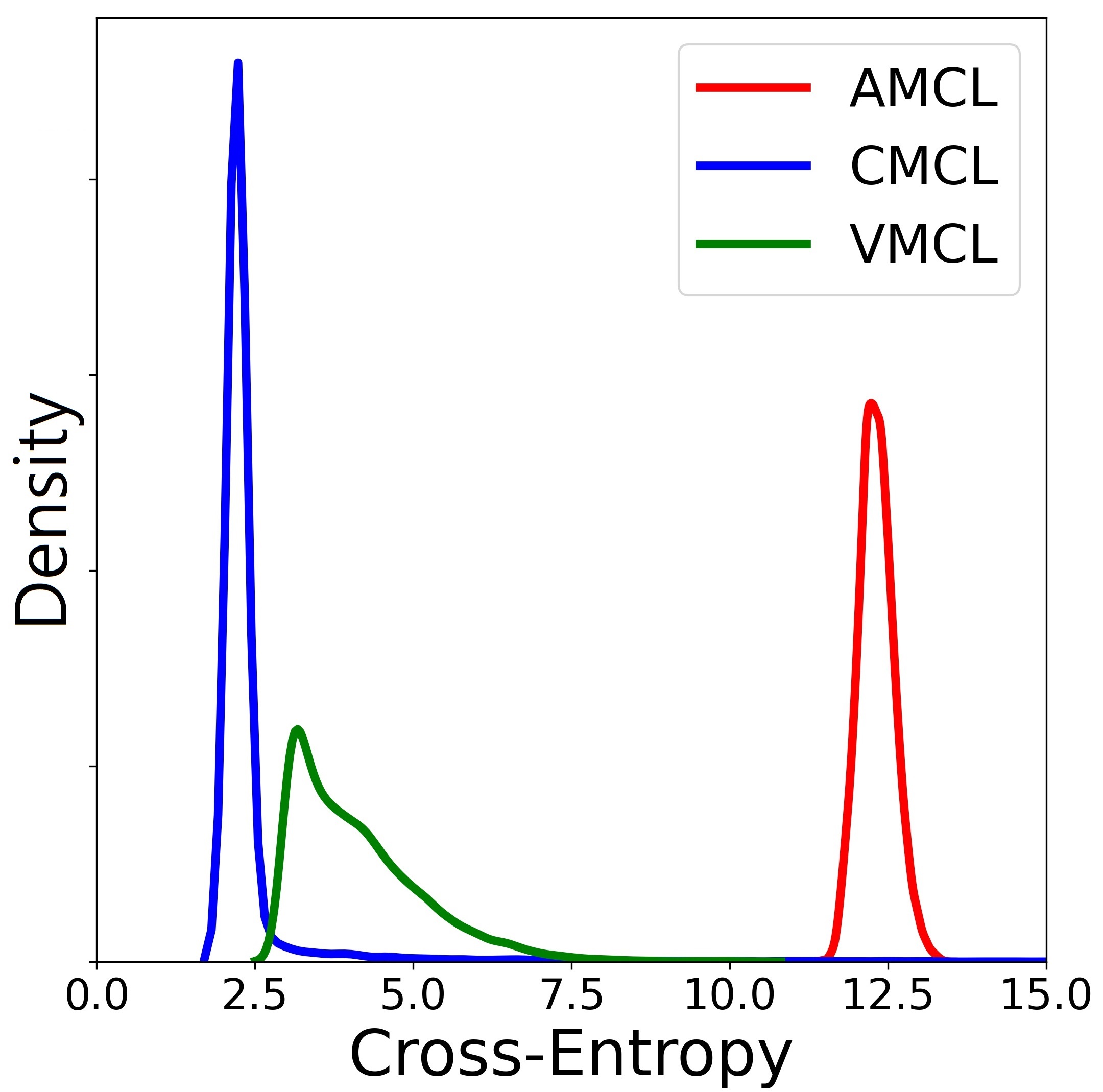}} 
	\caption{The distribution of measurements for analyzing the effect of auxiliary class. The classes of CIFAR-10 are separated into specialized and non-specialized by the ``Fixed specialization'' matrix generated during the training. (a) The probability density of auxiliary class on specialized, non-specialized, and SVHN (unseen) test data. (b)-(c) The cross-entropy of the test examples from (b) specialized and (c) non-specialized classes.}
	\label{fig:analysis}
\end{figure*}

Due to the auxiliary class, AMCL effectively prevents overconfident predictions to the non-specialized classes by setting the ground truth label to $\mathcal{A}(\widetilde{y})$(See Equation \eqref{eq:amcl_loss1} and \eqref{eq:amcl_loss2}).
Since the probability of auxiliary class explicitly indicates whether or not an input example is specialized to the corresponding model, it can be exploited to measure the uncertainty on unseen examples. 
To verify the quality of the uncertainty level through AMCL, we investigate the probability of the auxiliary class for the images from SVHN~\cite{svhn} using an ensemble model consists of 5 ResNet-18 trained by only the CIFAR-10 dataset with the overlap parameter $K=1$.
As presented in \figref{fig:ac_anal}, AMCL produces a high probability towards the auxiliary class on the unseen data, which is almost as high as the probability of the non-specialized classes used in training.
Additionally, we examine the cross-entropies between the ground truth labels and the predictions of the images from specialized and non-specialized classes, separately, as presented in \figrefs{fig:special_anal}{fig:nonspecial_anal}. 
The predictions are from the models trained with three variants of MCL. 
In contrast to the other methods, AMCL shows a large margin between the cross-entropies from the specialized and the non-specialized classes because AMCL distinctively knows the specialization of each model through the auxiliary class.
This cross-entropy results demonstrate that AMCL achieves the most extreme specialization compared to the previous MCL approaches.

\begin{table}[t!]
	\centering
	\small
	\begin{tabular}{c c c l l}\hline
		Method & MBA & FFM & Oracle & Top-1 \\ \hline
		IE  & - & - & 5.10 & 10.18 \\ 
		sMCL  & - & - & 3.27 & 56.54 \\ 
		CMCL  & - & - & 3.36 & 10.83 \\ 
		vMCL  & - & - & 3.54 & 11.49 \\ \hline
		\multirow{4}{*}{AMCL}& - & - & 3.42 & 12.42 \\ 
		& $\surd$ & - & 2.54 & 11.63 \\
		& - & $\surd$ & 3.13 & 10.86 \\ 
		& $\surd$ & $\surd$ & \tbf{2.14}\textsubscript{\tbf{(-36.3\%)}} & \tbf{9.74}\textsubscript{\tbf{(-10.1\%)}} \\ \hline
	\end{tabular}
	\caption{The classification error rates (\%) on CIFAR-10 with the IE and the previous MCL methods. 
		``MBA'' and ``FFM'' are short for memory-based assignment and feature fusion module, respectively. 
		``$\surd$'' implies selected. 
		The best score is marked in \textbf{bold}, and the values in parentheses as subscript represent the relative error reductions from CMCL, which is the second best result in terms of harmonic mean.}
	\label{table:ablation}
\end{table}

\subsubsection{Effect of feature fusion module}
We also examine the comparative performance of feature fusion module by replacing it with the corresponding techniques from CMCL and vMCL.
The dataset and the network used for this study are identical to those of the ablation study.
The feature fusion module is inserted just before the first pooling layer, which is the same to the application of feature sharing in CMCL.

\begin{figure}[t!]
	\centering
	\includegraphics[width=1\columnwidth, height=0.42\columnwidth]{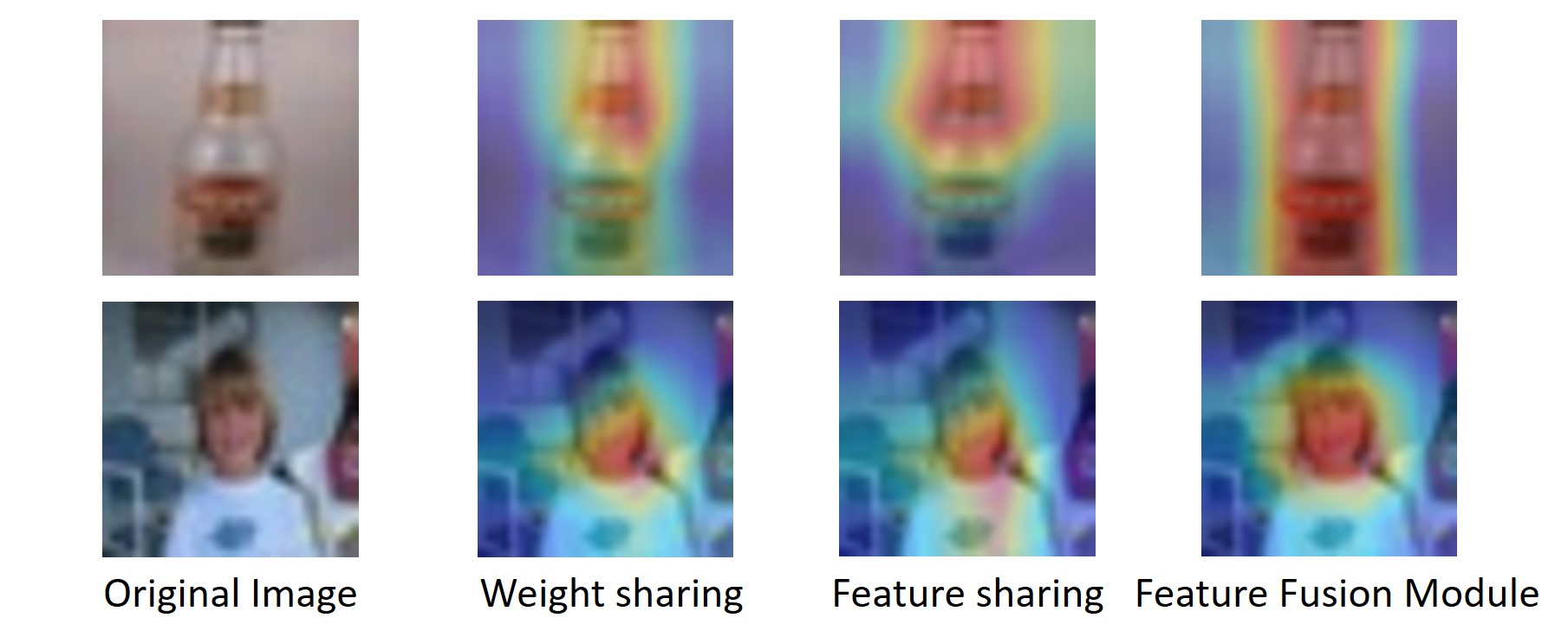}
	\caption{Grad-CAM visualizations of the feature maps from the specialized models trained with different generalization modules. Best viewed in color.}
	\label{fig:ffm}
\end{figure}

As shown in~\tabref{table:ffm}, the use of feature fusion module results in more improvement over the other techniques.
Although the number of total parameters in the network increases by 1.8$\%$, the improvement in classification, 16.7$\%$ and 5.8$\%$ relative error reductions in the oracle and the top-1 errors, shows that feature fusion module achieves better generalization.
Additionally for a qualitative analysis, the Grad-CAM visualizations~\cite{gradcam} are examined to check the regions of the input focused to make the predictions. 
As presented in~\figref{fig:ffm}, 
the feature maps from the model trained with our feature fusion module highlight more definitive regions of the objects. 
Additional visualizations from both specialized and non-specialized models are presented in the supplementary material.


\subsubsection{Comparison with MCL methods}
We now compare the performance using standard, large-scale convolutional networks on three public datasets.
As summarized in~\tabref{table:cnns_result}, AMCL consistently shows improvements compared to the other MCL methods.
Specifically, in the case with the CIFAR-10 dataset, our method with GoogLeNet shows the most impressive result producing $15.79\%$ and $6.25\%$ relative error reductions from the second best result in terms of the oracle and the top-1 errors, respectively.
Even though both CIFAR-100 and Tiny-ImageNet are more challenging datasets with significantly larger numbers of classes compared to CIFAR-10, AMCL shows the superiority with all types of CNNs.
Interestingly, we observe that only our method improves the top-1 error rate compared to the results of IE with Tiny-ImageNet, which none of the other methods have achieved.

\subsubsection{Comparison with non-MCL ensemble methods}
To show the reliability of AMCL, we compare the performance of the recent ensemble methods, including EMADNN~\cite{ennm}, ONE~\cite{one} and KDCL~\cite{kdcl}, on CIFAR-10/100.
EMADNN enforces the diversity by minimizing the similarity between all pairs of the base models, while ONE and KDCL apply the concept of knowledge distillation~\cite{hinton_kd} among the base models.
The main difference of our method compared to these previous works is to lead each base model to become a specialist for a certain subset of the dataset, not on the entire set.
In this comparision, we test PreAct ResNet-10~\cite{preact_resnet} and MobileNetV2~\cite{mobilenetv2} as the base models.
As presented in \tabref{table:other_ensemble}, AMCL outperforms by large margins in terms of oracle error rates and shows impressive top-1 error rates superior to those of all baselines in most cases.

\begin{table}[t!]
	\centering
	\small
	\begin{tabular}{c l l | l l}\hline
		\multirow{2}{*}{Method} & \multirow{2}{*}{Oracle} & \multirow{2}{*}{Top-1} & \multirow{1}{*}{Total} & \multirow{1}{*}{Relative} \\
		&  &  & \multirow{1}{*}{Parameters} & \multirow{1}{*}{Ratio}\\ \hline
		WS  & 2.87 & 10.91 & 55,458K & 1.000 \\ 
		FS  & 2.57 & 10.34 & 56,058K & 1.011 \\ \hline
		FFM & \tbf{2.14}\textsubscript{\tbf{(-16.7\%)}} & \tbf{9.74}\textsubscript{\tbf{(-5.8\%)}} & 56,444K & 1.018 \\ \hline
	\end{tabular}
	\caption{The classification error rates (\%) and the number of parameters with different generalization modules. 
		``WS'' is the weight sharing from vMCL, and ``FS'' indicates the feature sharing scheme from CMCL. ``FFM'' is feature fusion module of our method.  
		Auxiliary class and memory-based assignment are applied in all comparisons.
		The best score is marked in \textbf{bold}, and the values in parentheses as subscript represent the relative error reductions from FS, the second best result in terms of harmonic mean.}
	\label{table:ffm}
\end{table}

\begin{table*}[t!]
	\centering
	\small
	\begin{tabular}{c l | l l | l l | l l}
		\hline
		\multirow{1}{*}{Model} &
		\multirow{1}{*}{Ensemble} &
		\multicolumn{2}{c}{CIFAR-10} &
		\multicolumn{2}{c}{CIFAR-100} &
		\multicolumn{2}{c}{Tiny-ImageNet} \\ \cline{3-8}
		\multirow{1}{*}{Name} & \multirow{1}{*}{Method} & Oracle & Top-1 & Oracle & Top-1 & Oracle & Top-1 \\ \hline
		\multirow{5}{*}{VGGNet-16} &  IE  &  3.44$_{\pm \tx{0.04}}$ & 8.26$_{\pm \tx{0.12}}$ & 17.78$_{\pm \tx{0.11}}$ & 31.08$_{\pm \tx{1.13}}$ & 27.42$_{\pm \tx{0.46}}$ & \blue{40.63$_{\pm \tx{0.18}}$} \\ 
		& sMCL &  1.98$_{\pm \tx{0.69}}$ & 21.29$_{\pm \tx{1.40}}$ & \blue{15.97$_{\pm \tx{0.13}}$} & 32.06$_{\pm \tx{0.38}}$ & 26.68$_{\pm \tx{0.22}}$ & 41.60$_{\pm \tx{0.33}}$ \\
		& CMCL &  2.70$_{\pm \tx{0.12}}$ & \blue{7.72$_{\pm \tx{0.24}}$} & 16.63$_{\pm \tx{0.23}}$ & \blue{29.29$_{\pm \tx{0.29}}$} & \blue{26.11$_{\pm \tx{0.28}}$} & 40.88$_{\pm \tx{0.17}}$ \\
		& vMCL &  1.75$_{\pm \tx{0.09}}$ & 11.43$_{\pm \tx{0.34}}$ & 24.19$_{\pm \tx{0.67}}$ & 43.77$_{\pm \tx{1.15}}$ & 39.05$_{\pm \tx{0.48}}$ & 56.02$_{\pm \tx{1.07}}$ \\
		& AMCL &  \tbf{1.30}$_{\pm \tbf{0.12}}$ & \tbf{7.24}$_{\pm \tbf{0.13}}$ & \tbf{14.79}$_{\pm \tbf{0.15}}$ & \tbf{28.58}$_{\pm \tbf{0.22}}$ & \tbf{24.24}$_{\pm \tbf{0.15}}$ & \tbf{38.59}$_{\pm \tbf{0.21}}$ \\ \hline
		\multirow{5}{*}{ResNet-34} & IE   & 4.64$_{\pm \tx{0.12}}$ & 10.00$_{\pm \tx{0.20}}$ & 21.23$_{\pm \tx{0.11}}$ & 34.47$_{\pm \tx{0.18}}$ & 25.62$_{\pm \tx{0.36}}$ & \blue{38.03$_{\pm \tx{0.26}}$} \\ 
		& sMCL & 2.95$_{\pm \tx{0.17}}$ & 13.43$_{\pm \tx{0.18}}$ & 22.44$_{\pm \tx{0.32}}$ & 39.58$_{\pm \tx{0.15}}$ & 26.12$_{\pm \tx{0.37}}$ & 40.91$_{\pm \tx{0.38}}$ \\
		& CMCL & \blue{2.22$_{\pm \tx{0.19}}$} & \blue{8.13$_{\pm \tx{0.20}}$} & \blue{16.09$_{\pm \tx{0.32}}$} & \blue{29.73$_{\pm \tx{0.33}}$} & \blue{24.60$_{\pm \tx{0.33}}$} & 39.17$_{\pm \tx{0.45}}$ \\
		& vMCL & 2.63$_{\pm \tx{0.11}}$ & 12.05$_{\pm \tx{0.18}}$ & 19.21$_{\pm \tx{0.36}}$ & 33.35$_{\pm \tx{0.17}}$ & 26.80$_{\pm \tx{0.38}}$ & 39.42$_{\pm \tx{0.83}}$ \\
		& AMCL & \tbf{2.09}$_{\pm \tbf{0.10}}$ & \tbf{7.42}$_{\pm \tbf{0.11}}$ & \tbf{14.76}$_{\pm \tbf{0.24}}$ & \tbf{27.94}$_{\pm \tbf{0.21}}$ & \tbf{22.56}$_{\pm \tbf{0.37}}$ & \tbf{36.94}$_{\pm \tbf{0.31}}$ \\ \hline
		\multirow{5}{*}{GoogLeNet} & IE   & 3.22$_{\pm \tx{0.05}}$ & 7.53$_{\pm \tx{0.08}}$ & 14.99$_{\pm \tx{0.29}}$ & \blue{25.65$_{\pm \tx{0.14}}$} & 21.49$_{\pm \tx{0.35}}$ & \blue{32.77$_{\pm \tx{0.20}}$} \\ 
		& sMCL & 2.31$_{\pm \tx{0.11}}$ & 9.72$_{\pm \tx{0.19}}$ & \blue{13.40$_{\pm \tx{0.17}}$} & 28.36$_{\pm \tx{0.29}}$ & 21.38$_{\pm \tx{0.36}}$ & 34.67$_{\pm \tx{0.30}}$ \\
		& CMCL & 2.28$_{\pm \tx{0.08}}$ & \blue{7.52$_{\pm \tx{0.26}}$} & 13.51$_{\pm \tx{0.37}}$ & 25.97$_{\pm \tx{0.34}}$ & \blue{21.38$_{\pm \tx{0.47}}$} & 33.06$_{\pm \tx{0.42}}$ \\
		& vMCL & \blue{2.09$_{\pm \tx{0.18}}$} & 8.67$_{\pm \tx{0.32}}$ & 14.11$_{\pm \tx{0.28}}$ & 26.67$_{\pm \tx{0.46}}$ & 23.26$_{\pm \tx{0.67}}$ & 33.07$_{\pm \tx{0.53}}$ \\
		& AMCL & \tbf{1.76}$_{\pm \tbf{0.10}}$ & \tbf{7.05}$_{\pm \tbf{0.22}}$ & \tbf{12.86}$_{\pm \tbf{0.09}}$ & \tbf{24.42}$_{\pm \tbf{0.26}}$ & \tbf{20.24}$_{\pm \tbf{0.32}}$ & \tbf{31.65}$_{\pm \tbf{0.28}}$ \\ \hline
	\end{tabular} 
	\caption{The comprehensive classification results with various CNNs. 
		We report the mean and standard deviation~(as subscript) of the oracle and the top-1 error rates(\%) by repeating 3 times.
		Among the results with the overlap parameter $K=1,\cdots,4$, we pick the best result in terms of harmonic mean.
		The best score is marked in \textbf{bold}, and the second best result is in blue.}
	\label{table:cnns_result}
\end{table*}

\begin{table*}[t!]
	\centering
	\small
	\begin{tabular}{c l | l l | l l}
		\hline
		\multirow{2}{*}{Base Models} & \multirow{1}{*}{Ensemble} & \multicolumn{2}{c}{CIFAR-10} & \multicolumn{2}{c}{CIFAR-100} \\ \cline{3-6}
		& \multirow{1}{*}{Method} & Oracle & Top-1 & Oracle & Top-1 \\ \hline		
		& IE   & 7.60$_{\pm \tx{0.70}}$ & 12.12$_{\pm \tx{0.12}}$ & 28.87$_{\pm \tx{0.34}}$ & 41.79$_{\pm \tx{0.12}}$  \\ 
		PreAct & EMADNN & 7.36$_{\pm \tx{0.64}}$ & 11.84$_{\pm \tx{0.10}}$ & 31.32$_{\pm \tx{0.59}}$ & 41.22$_{\pm \tx{0.83}}$  \\ 
		ResNet-10 & ONE  & 8.76$_{\pm \tx{0.21}}$ & \blue{11.30$_{\pm \tx{0.11}}$} & 33.88$_{\pm \tx{0.27}}$ & 40.76$_{\pm \tx{0.15}}$  \\ 
		& KDCL & \blue{7.29$_{\pm \tx{0.22}}$} & 11.91$_{\pm \tx{0.08}}$ & \blue{28.83$_{\pm \tx{0.11}}$} & \blue{39.89$_{\pm \tx{0.27}}$}  \\ 
		& AMCL & \tbf{4.88}$_{\pm \tbf{0.04}}$~\textsubscript{\tbf{(-33.1\%)}} & \tbf{10.78}$_{\pm \tbf{0.07}}$~\textsubscript{\tbf{(-4.6\%)}} & \tbf{27.45}$_{\pm \tbf{0.48}}$~\textsubscript{\tbf{(-4.8\%)}} & \tbf{37.91}$_{\pm \tbf{0.52}}$~\textsubscript{\tbf{(-5.0\%)}} \\ \hline
		
		\multirow{5}{*}{MobileNetV2} & IE  & \blue{5.60$_{\pm \tx{0.51}}$} & 10.25$_{\pm \tx{0.20}}$ & 22.48$_{\pm \tx{0.45}}$ & 31.87$_{\pm \tx{0.28}}$  \\ 
		& EMADNN & 6.84$_{\pm \tx{0.33}}$ & 10.53$_{\pm \tx{0.72}}$ & 29.68$_{\pm \tx{0.30}}$ & 35.91$_{\pm \tx{0.30}}$  \\ 
		& ONE  & 7.31$_{\pm \tx{0.06}}$ & \blue{10.19$_{\pm \tx{0.09}}$} & 23.48$_{\pm \tx{0.08}}$ & 31.89$_{\pm \tx{0.11}}$  \\ 
		& KDCL & 6.22$_{\pm \tx{0.16}}$ & 10.90$_{\pm \tx{0.08}}$ & \blue{22.41$_{\pm \tx{0.12}}$} & \tbf{30.08}$_{\pm \tbf{0.40}}$  \\ 
		& AMCL & \tbf{3.99}$_{\pm \tbf{0.18}}$~\textsubscript{\tbf{(-28.8\%)}} & \tbf{9.29}$_{\pm \tbf{0.13}}$~\textsubscript{\tbf{(-8.8\%)}} & \tbf{21.79}$_{\pm \tbf{0.17}}$~\textsubscript{\tbf{(-2.8\%)}} & 30.56$_{\pm \tx{0.17}}$ \\ \hline
	\end{tabular} 
	\caption{The classification error rates~(\%) on CIFAR-10/100 with various ensemble methods. All methods are used to train an ensemble of 3 networks. EMADNN uses 16 multi-branch networks, and KDCL is trained with KDCL-Naive for the ensemble logits. AMCL is trained with the overlap parameter $K=2$. We report the mean and standard deviation~(as subscript) by repeating each test 3 times. The best score is marked in \textbf{bold}, and the values in parentheses as subscript represent the relative reductions from the second best result in blue.}
	\label{table:other_ensemble}
\end{table*}

\begin{table*}[t!]
	\centering
	\small
	\begin{tabular}{c l l | l l | l l | l l}
		\hline
		Method &
		\multicolumn{2}{c}{sMCL} &
		\multicolumn{2}{c}{CMCL} &
		\multicolumn{2}{c}{vMCL} &		
		\multicolumn{2}{c}{AMCL} \\ \hline
		M & Oracle & Top-1 & Oracle & Top-1 & Oracle & Top-1 & Oracle & Top-1 \\ \hline		
		1 &  15.05 & 15.05 & 15.05 & 15.05 & 15.05 & 15.05 & 15.05 & 15.05 \\ 
		2 &  11.71 & 15.12 & 13.33 & \blue{13.31} & \blue{10.73} & 15.49 & \tbf{10.48}\textsubscript{\tbf{(-2.3\%)}} & \tbf{11.53}\textsubscript{\tbf{(-13.4\%)}} \\
		3 &  10.65 & 12.67 & 10.86 & \blue{11.20} & \blue{10.27} & 12.26 & \tbf{8.68}\textsubscript{\tbf{(-15.5\%)}} & \tbf{9.60}\textsubscript{\tbf{(-14.3\%)}} \\
		4 &  9.52 & 13.25 & 10.42 & \blue{11.19} & \blue{9.30} & 12.88 & \tbf{7.97}\textsubscript{\tbf{(-14.3\%)}} & \tbf{8.83}\textsubscript{\tbf{(-21.1\%)}} \\
		5 &  \blue{9.37} & 11.82 & 9.75 & \blue{10.58} & 9.44 & 12.06 & \tbf{7.66}\textsubscript{\tbf{(-18.2\%)}} & \tbf{8.61}\textsubscript{\tbf{(-18.6\%)}} \\ \hline
	\end{tabular} 
 	\caption{The foreground-background segmentation error rates (\%) on the iCoseg dataset with varying values of the ensemble size $M$.
	We report the mean by repeating 3 times without the standard deviation due to the space limitation.
	The best score is marked in \textbf{bold}, and the values in parentheses as subscript represent the relative reductions from the second best result in blue.}
	\label{table:seg_result}	
\end{table*}

\subsection{Image segmentation}
In this section, we evaluate to verify that AMCL produces higher performance in the image segmentation task using the iCoseg dataset. 
To evaluate the image segmentation task, we define the prediction error rate as the percentage of incorrectly labeled pixels.
The oracle error rate is defined by the lowest prediction error among the ensemble models. 
Meanwhile, the top-1 error is measured by the error rate of the final prediction.
For a fair comparison, we report both the oracle and the top-1 errors with varying ensemble sizes. 
\tabref{table:seg_result} shows that AMCL significantly outperforms all ensemble methods.
With 5 model ensemble, AMCL provides $18.2\%$ and $18.6\%$ relative reductions from the second best results in the oracle and the top-1 errors, respectively.
These experiments demonstrate the reliability of AMCL in image segmentation beyond classification.
The network architecture and the experiment results with more elaborated visualizations are illustrated in the supplementary material.

\section{Conclusion}
This paper presents an advanced ensemble method to train for an extreme model specialization which results in diverse solutions of high quality.
For this purpose, we introduce simple but effective techniques: 1) auxiliary class, 2) memory-based assignment, and 3) feature fusion module.
The extensive experiments with various networks and large datasets show that AMCL results in better performances in most cases. 
We believe that our newly proposed approach can be applied in many related application fields with slight modifications and provide better results.

\bibliography{amclbib}

\end{document}


\pagestyle{empty} 


\centerline{{\huge \textbf{ Supplementary Material:}}}
\vspace{5mm}
\centerline{{\LARGE \textbf{Auxiliary Class Based Multiple Choice Learning}}}

\vskip 0.25in




\begin{flushleft}
{\Large \textbf{A. Pseudo Code for AMCL}}
\end{flushleft}
We describe the full training procedure of AMCL with the pseudo code presented in Algorithm~\ref{alg:amcl_alg}. 
As explained in Section 3.2, the assignments of model specialization and the corresponding computation of loss change after a set number of epochs, $T_{\tau}$. When the number of epoch is smaller than $T_{\tau}$, an input example is assigned to the models with the lower losses. However, after $T_{\tau}$, the assignment is fixed by the counts from the prior iterations, and the gradient of the loss is directly computed.  

\begin{algorithm}[ht!]
	\small
	\SetAlgoLined
	\KwIn{Training dataset $\mathcal{D} = \{\left(\bx_j,y_j\right)_{j=1}^N\}$, parameters $\{\beta, \gamma\}$ and training threshold $T_{\tau}$ }
	\KwOut{Trained $M$ ensemble models}
	\hrulefill
	
	Let $\mathcal{A}(\widetilde{y})$ be a one-hot encoded vector where only the auxiliary class is set to 1 and $N_c$ denotes the total number of classes \\
	Initialize $\{\theta_m\}_{m=1}^M$ and set $w_{k}^{m} = 0,~\forall k,m$ \\
	\Repeat{convergence}{
		Sample a batch $\mathcal{B} \subset D$ \\
		\uIf{$t_s \leq T_{\tau}$ } {
			$/ \ast$ Compute the loss for each model \\
			$L_j^m \leftarrow \ell\Big( \widetilde{y}_j, P_{\theta_m}(\widetilde{y}~|~\bx_j) \Big) + \beta\dsum_{\hat{m} \neq m} D_{KL}\Big( \mathcal{A}(\widetilde{y})~||~P_{\theta_{\hat{m}}}(\widetilde{y}~|~\bx_j) \Big) \quad \forall (\bx_j, y_j) \in \mathcal{B}$ \\ 
			\For{$j = 1$ \KwTo $|\mathcal{B}|$}{
				Set $v_j^m=0, \quad m=1,\cdots,M$ \\
				$\overline{m} \leftarrow $ Select $K$ models with the lowest $L_j^m $ \\
				\For{$k = 1$ \KwTo $K$}{
				$v_j^{\overline{m}(k)} = 1 \qquad\qquad\qquad\quad ~~~/ \ast$ Set the indicator variable \\
				$w_{c(j)}^{\overline{m}(k)} \mathrel{+}= 1 \qquad\qquad\quad ~~~~~~~/ \ast$ Count the indicator variable for $\overline{m}(k)$-th model \\
				}
				$\partial \mathcal{L}_{loss\eh based}(\bx_j; \beta)/\partial \theta_m \qquad\quad ~/ \ast$ Compute the gradient with respect to $\theta_m,~\forall m$ \\
			}
		} \uElse{
			$/ \ast$ Binarize top-$K$ models for each class once at $t_s = T_{\tau}+1$ \\
			\For{$k = 1$ \KwTo $N_c$}{
				Binarize $[w_{k}^{1}, \cdots, w_{k}^{M}]$ to top-$K$
			}
			\For{$j = 1$ \KwTo $|\mathcal{B}|$}{
				$\partial \mathcal{L}_{memory\eh based}(\bx_j; \gamma)/\partial \theta_m \quad ~~~/ \ast$ Compute the gradient with respect to $\theta_m,~\forall m$ \\
			}
		}
		Update the model parameters
	}
	\caption{Auxiliary class based Multiple Choice Learning (AMCL)}
	\label{alg:amcl_alg}
\end{algorithm}

\begin{flushleft}
{\Large \textbf{B. Experimental Details}}
\end{flushleft}
In this section, we provide detailed setups used in our experiments.

\begin{flushleft}
{\large \textbf{B.1 Datasets}} 
\end{flushleft}
\vspace{-0.5em}
\begin{itemize}
	\item \textbf{SVHN} is a set of real-world number images(0 through 9) with 73,257 training and 26,032 testing examples. Each image has 32 $\times$ 32 RGB pixels.
	\item \textbf{CIFAR-10} is a set of object images from 10 different classes consisting of 50,000 training and 10,000 testing examples. 
	Each image has 32 $\times$ 32 RGB pixels.
	\item \textbf{CIFAR-100} is a dataset similar to CIFAR-10 with 100 classes instead of 10.
	\item \textbf{Tiny-ImageNet} is a subset of the ILSVRC 2012 classification dataset. It has 200 classes, and for each class, 500 training, 50 validation, and 50 test images are provided. All images are in 64 $\times$ 64 RGB pixels. 
	\item \textbf{iCoseg} consists of 643 images in 38 groups with the pixel-level ground truths on the foreground-background segmentation. It can be considered as pixel-level two-class classification problem, \ie, 0 (background) or 1 (foreground).
\end{itemize}

\begin{flushleft}
{\large \textbf{B.2 Implementations}} 
\end{flushleft}
\vspace{-0.5em}
All models in our experiments are trained from scratch. 
For both classification and segmentation tasks, we pre-process the images with global contrast normalization and do not augment any data in all experiments.
We train all models with the initial learning rate of 0.1 and drop it by 0.2 for every 25 epochs. 
The Nesterov momentum is set to 0.9 for SGD, and weight decay is set to 0.0005. 
The minibatch size is 128, and $T_\tau$, the epoch threshold used in memory-based assignment, is set to 5. 
Due to the inherent complexity of the data, we set the total epochs differently for each task.
In classification, the total number of epochs is set to 100, 125 and 150 for CIFAR-10, CIFAR-100 and TinyImageNet, respectively. 
Meanwhile, the total number of epochs is set to 300 in the segmentation task.

In segmentation task, following the experimental setting of CMCL, we select the images that are larger than $300 \times 500$ pixels and randomly split into the training and testing sets with the ratio of 8:2 from each class. 
The images are resized into $75 \times 125$ using bicubic interpolation to train the network based on a Fully Convolutional Network (FCN)~\cite{fcn} with the decoder architecture presented in~\cite{unsupervised2015}.
We train the FCN from scratch for all methods in comparison.

\newpage 
\begin{flushleft}
{\Large \textbf{C. Effects of Memory-based Assignment and Feature Fusion Module}} 
\end{flushleft}

\begin{figure*}[ht!]
	\centering
	\includegraphics[width=0.95\columnwidth, height=0.5\columnwidth]{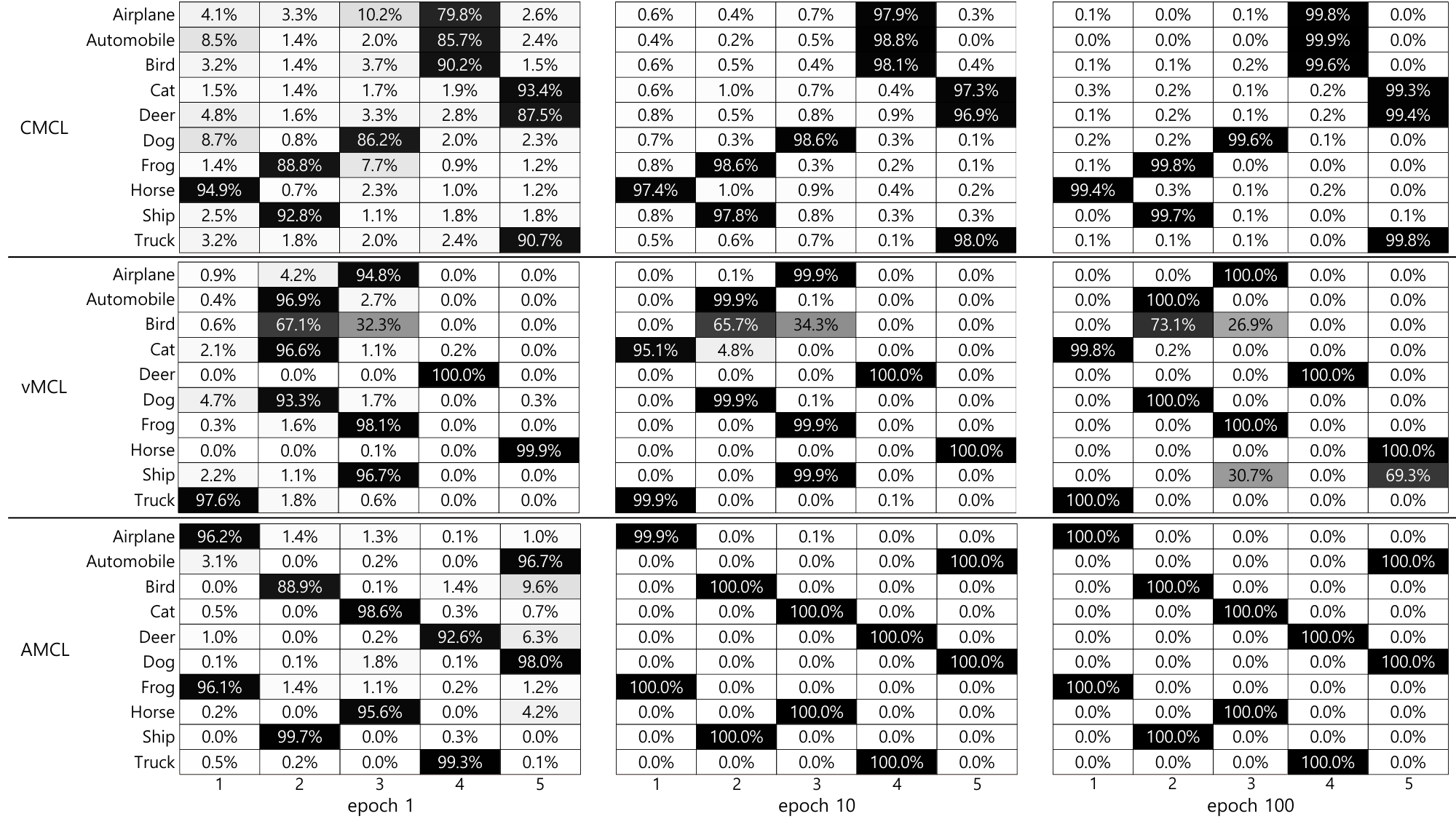}
	\caption{The change in within-class ratios for the cumulative sum of $v_j^m$ with $K=1$ at epoch 1, 10, and 100. As the training lasts, AMCL with the memory-based assignment shows more intensive ratios towards the specialized classes for each model even with respect to $v_j^m$, which is the indicator variable from the loss-based assignment (Please refer to Equation (5) in Section 3.2).} 
	\label{fig:la_flow}
\end{figure*}

\begin{flushleft}
{\large \textbf{C.1 Assignment of the Model Specialization with Different MCL methods}}
\end{flushleft}
\vspace{-0.5em}
Following from Section 3.2, we examine how the actual assignment of input examples to the models realize as the training epochs proceed. 
The ensembles with simple CNN models (See Section 3.1) are trained by CMCL, vMCL, and AMCL with the epoch threshold $T_\tau$ at 10.
As shown in \figref{fig:la_flow}, when the epoch is 10, about 2$\sim$3~$\%$ of the inputs from each class are not assigned properly to the specialized models using CMCL.
From the perspective of specialization, these inputs become potential noise, which indeed affect the performance.
In the case of AMCL, each model becomes a specialist for certain classes with high purity compared to the other methods.

\begin{figure*}[ht!]
	\centering
	\includegraphics[width=0.8\columnwidth, height=0.85\columnwidth]{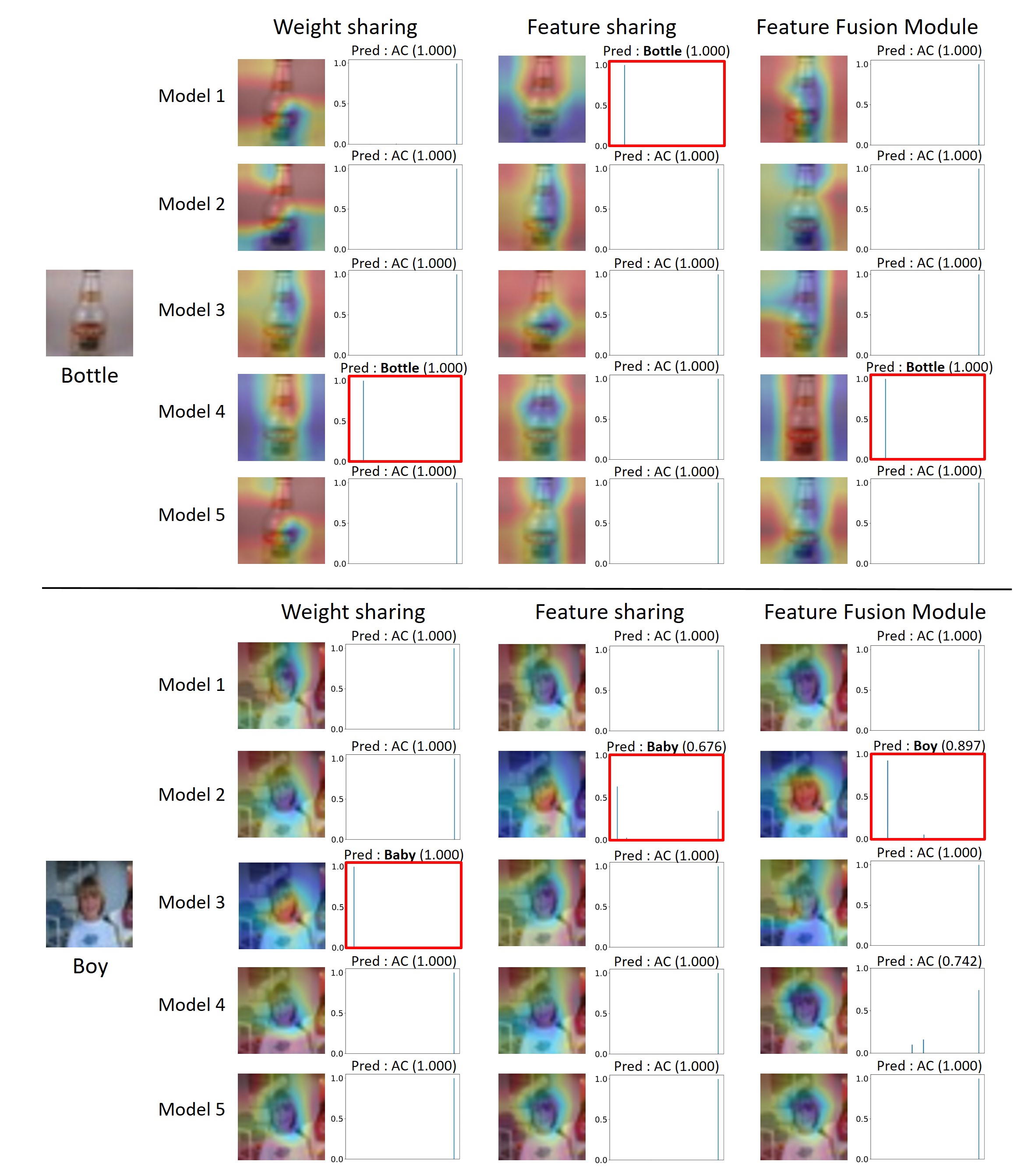}
	\caption{Grad-CAM visualizations of the feature maps and the predictive probabilities from all models in the ensembles of 5 ResNet-34 on CIFAR-100. The red boxes indicate the models specialized to the given classes after the training is completed. Since the overlap parameter $K$ is equal to 1, only one model is specialized to each class of images. Figure best viewed in color.} 
	\label{fig:gradcam}
\end{figure*}

\begin{flushleft}
{\large \textbf{C.2 Grad-CAM Visualizations with Different Generalization Modules}}
\end{flushleft}
\vspace{-0.5em}
The quantitative comparisons of our technique, \textit{feature fusion module}, abbreviated as FFM, to the other generalization modules, which are used in the previous MCL variants, are presented in Section 4.1. 
Although the changes in the classification error rates represent how much the FFM improves the performance compared to the other techniques, we additionally present the Grad-CAM visualizations~\cite{gradcam} of the feature maps from the models trained with different generalization modules in order to check how well the model trained with FFM focuses on the discriminative regions. 
From both examples presented in~\figref{fig:gradcam}, it is apparent that the model trained with FFM has better focuses on the objects themselves. Especially for the case with the boy image, presented on the bottom part of the figure, the models trained with weight sharing and feature sharing do not fully concentrate on the head of the boy. These two ensembles wrongly predict the boy example to a baby.  

\newpage 
\begin{flushleft}
{\Large \textbf{D. In-depth Studies on the Training and the Hyperparameters}} 
\end{flushleft}

\begin{flushleft}
{\large \textbf{D.1 Penalty Parameters $\beta$ and $\gamma$}} 
\end{flushleft}
\vspace{-0.5em}
We investigate the performance with varying $\beta$ and $\gamma$ values to check the sensitivity.
\tabref{table:hyper_result} shows the changes in the error rates with the combinations of $\beta \in \{0.01,0.05,0.1,0.2,0.5\}$ and $\gamma \in \{0.01,0.05,0.1,0.5,1.0,2.0\}$ on the CIFAR-10 dataset.
From the results, we can point out that the performance of AMCL with respect to the oracle error decreases when the value of $\beta$ gets larger. 
In terms of the optimization process, a larger $\beta$ value imposes more focus on leading the predictions of the non-specialized models to $\mathcal{A}(\widetilde{y})$ in the early iterations rather than minimizing the loss of the specialized models (Please refer to Equation (5) in Section 3.2).
As this case gets extreme, some models get specialized to all the images while the others do not get specialized to any.
Therefore, increasing $\beta$ naturally leads to the polarization of model specialization which is against the idea of diversifying the base models in the ensemble and results in high oracle errors.
However, even when $\beta$ increases, the performance with respect to the top-1 error does not show an explicit trend since the top-1 error is measured by averaging the predictions from all models, including the weak predictions from the non-specialized models.
In contrast to the results with $\beta$, we could not find an apparent trend in the performance with respect to the value of $\gamma$. 
Through this study, we set the hyperparameters $\beta = 0.01$ and $\gamma = 0.5$ in all the other experiments.

\begin{table}[ht]
	\centering
	\smallskip
	\begin{tabular}{c|c c c c c c }\hline
		\diagbox[innerleftsep=1em,width = 7em]{$\beta$}{$\gamma$} & 0.01 & 0.05 & 0.1 & 0.5 & 1.0 & 2.0 \\ \hline
		0.01 & 2.84~/~14.31  & 2.33~/~12.83 & 2.46~/~12.26  & \tbf{2.38~/~11.62} & 2.43~/~12.05 & 2.51~/~13.63  \\
		0.05 & 2.33~/~14.10  & 2.65~/~12.23 & 2.42~/~12.86  & 3.07~/~11.97 & 2.25~/~12.12 & 3.24~/~13.23 \\
		0.1  & 3.53~/~14.39  & 3.89~/~13.12 & 2.60~/~12.72  & 3.32~/~12.60 & 3.23~/~12.75 & 4.10~/~13.80 \\
		0.2  & 5.06~/~12.18  & 5.94~/~12.31 & 5.22~/~12.79  & 5.32~/~12.81 & 5.95~/~11.90 & 5.62~/~12.80 \\
		0.5  & 5.89~/~12.52  & 5.27~/~12.26 & 5.22~/~12.10  & 5.81~/~12.36 & 6.45~/~12.07 & 6.43~/~12.90 \\ \hline
	\end{tabular}
	\caption{The classification error rates~(\%) of an ensemble of 5 ResNet-10 on CIFAR-10 with the combinations of hyperparameters $\beta$ and $\gamma$. The overlap parameter $K$ is set to 1. The results are represented as oracle~/~top-1. The best results, in terms of harmonic mean, are marked in \textbf{bold}.}
	\label{table:hyper_result}
\end{table}

\begin{flushleft}
{\large \textbf{D.2 Overlap Parameter $K$}}
\end{flushleft}
\vspace{-0.5em}
Similar to other MCL methods, AMCL can pick the top-$K$ specialized models instead of only one at the training stage.
Overlapping implies that there exist $K$ specialized models for an individual example, \ie, $\sum_{m=1}^M v_j^m = K$.
As $K$ increases, the weak generalization issue can be naturally resolved since the data capacity of each model increases. 
Hence, all ensemble methods including AMCL tend to show better performance in the top-1 error as $K$ increases.

\tabref{table:cnn_result_k} compares the performance on all datasets with an ensemble of 5 GoogLeNet.
AMCL consistently outperforms all baselines with respect to the oracle and top-1 error rates.	
As shown throughout the paper, finding the right overlap parameter $K$ affects the overall performance.
However, when $K$ gets close to $M$, all MCL methods becomes the same as the independent ensemble model, and thus, $K$, samller than $M$, have to be searched thoroughly. 

\begin{table*}[ht!]
	\centering
	\medskip
	\begin{tabular}{c c l l | l l | l l }
		\hline
		\multirow{1}{*}{Ensemble} &
		\multirow{2}{*}{$K$} &
		\multicolumn{2}{c}{CIFAR-10} &
		\multicolumn{2}{c}{CIFAR-100} &
		\multicolumn{2}{c}{Tiny-ImageNet} \\ \cline{3-8}
		\multirow{1}{*}{Method} & & Oracle  & Top-1 & Oracle & Top-1 & Oracle & Top-1 \\ \hline
		IE & - & 3.22$_{\pm 0.15}$ & 7.53$_{\pm 0.18}$ & 14.99$_{\pm 0.29}$ & 25.65$_{\pm 0.14}$ & 21.49$_{\pm 0.35}$ & 32.77$_{\pm 0.20}$ \\ \cline{1-8}
		\multirow{4}{*}{sMCL} & 1 & 2.16$_{\pm 0.11}$ & 50.90$_{\pm 0.33}$ & 19.28$_{\pm 0.12}$ & 45.61$_{\pm 0.23}$ & 31.67$_{\pm 0.15}$ & 54.21$_{\pm 0.21}$ \\
		& 2 & 2.08$_{\pm 0.10}$ & 19.77$_{\pm 0.74}$& 14.58$_{\pm 0.27}$ & 32.56$_{\pm 0.21}$ & 24.87$_{\pm 0.31}$ & 42.72$_{\pm 0.32}$ \\ 
		& 3 & 2.31$_{\pm 0.11}$ & 9.72$_{\pm 0.19}$ & 13.40$_{\pm 0.17}$ & 28.36$_{\pm 0.29}$ & 22.23$_{\pm 0.20}$ & 36.96$_{\pm 0.25}$ \\ 
		& 4 & 2.69$_{\pm 0.17}$	& 8.26$_{\pm 0.08}$ & 13.83$_{\pm 0.31}$ & 26.67$_{\pm 0.38}$ & 21.38$_{\pm 0.36}$ & 34.67$_{\pm 0.31}$ \\ \cline{1-8}
		\multirow{4}{*}{CMCL} & 1 & 4.57$_{\pm 0.28}$	& 10.08$_{\pm 0.15}$& 23.64$_{\pm 0.12}$ & 34.55$_{\pm 0.14}$ & 36.90$_{\pm 0.48}$ & 50.43$_{\pm 0.41}$ \\
		& 2 & 2.77$_{\pm 0.28}$	& 8.03$_{\pm 0.29}$	& 15.55$_{\pm 0.54}$ & 29.39$_{\pm 0.52}$ & 29.05$_{\pm 0.28}$ & 43.58$_{\pm 0.46}$ \\ 
		& 3 & 2.28$_{\pm 0.08}$	& 7.52$_{\pm 0.26}$	& 13.67$_{\pm 0.10}$ & 26.94$_{\pm 0.19}$ & 22.99$_{\pm 0.33}$ & 33.39$_{\pm 0.39}$ \\ 
		& 4 & 2.33$_{\pm 0.09}$	& 7.23$_{\pm 0.15}$	& 13.51$_{\pm 0.37}$ & 25.97$_{\pm 0.34}$ & 21.38$_{\pm 0.77}$ & 33.06$_{\pm 0.43}$ \\ \cline{1-8}
		\multirow{4}{*}{vMCL} & 1 & 4.33$_{\pm 0.23}$	& 11.61$_{\pm 0.39}$ & 22.65$_{\pm 0.39}$ & 36.57$_{\pm 0.38}$ & 36.55$_{\pm 0.66}$ & 52.23$_{\pm 0.57}$ \\
		& 2 & 3.24$_{\pm 0.26}$	& 9.24$_{\pm 0.21}$ & 15.91$_{\pm 0.26}$ & 32.82$_{\pm 0.41}$ & 29.46$_{\pm 0.53}$ & 44.43$_{\pm 0.60}$ \\ 
		& 3 & 2.09$_{\pm 0.18}$	& 8.67$_{\pm 0.62}$ & 14.68$_{\pm 0.36}$ & 28.36$_{\pm 0.49}$ & 25.07$_{\pm 0.71}$ & 34.92$_{\pm 0.50}$ \\ 
		& 4 & 3.01$_{\pm 0.46}$	& 8.23$_{\pm 0.24}$ & 14.11$_{\pm 0.28}$ & 26.67$_{\pm 0.47}$ & 23.26$_{\pm 0.87}$ & 33.07$_{\pm 0.84}$ \\ \cline{1-8}
		\multirow{4}{*}{AMCL} & 1 & 3.42$_{\pm 0.12}$	& 8.73$_{\pm 0.17}$	& 17.81$_{\pm 0.11}$ & 33.71$_{\pm 0.23}$ & 32.14$_{\pm 0.29}$ & 50.96$_{\pm 0.22}$ \\
		& 2 & 2.31$_{\pm 0.07}$ & 7.65$_{\pm 0.35}$ & 13.33$_{\pm 0.26}$ & 28.99$_{\pm 0.22}$ & 25.16$_{\pm 0.33}$ & 42.61$_{\pm 0.30}$ \\ 
		& 3 & \tbf{1.76}$_{\pm \tbf{0.10}}$	& \tbf{7.05}$_{\pm \tbf{0.22}}$ & \tbf{12.33}$_{\pm \tbf{0.23}}$ & 26.25$_{\pm 0.18}$ & 22.46$_{\pm 0.20}$ & 34.02$_{\pm 0.39}$ \\ 
		& 4 & 2.28$_{\pm 0.08}$	& 7.09$_{\pm 0.17}$ & 12.86$_{\pm 0.09}$ & \tbf{24.42}$_{\pm \tbf{0.26}}$ & \tbf{20.24}$_{\pm \tbf{0.32}}$ & \tbf{31.65}$_{\pm \tbf{0.28}}$ \\ \hline
	\end{tabular}
	\caption{The classification error rates (\%) on all datasets with varying values of the overlap parameter $K$. We report the mean and standard deviation~(as subscript) by repeated 3 times, and the best results are marked in \textbf{bold}.}
	\label{table:cnn_result_k}
\end{table*}

\newpage
\begin{flushleft}
{\large \textbf{D.3 Extension of the Ensemble Size $M$}}
\end{flushleft}
\vspace{-0.5em}
We additionally show the superiority of AMCL with a larger ensemble size $M$.
Specifically, we train the ensemble model with 10 ResNet-10 using the CIFAR-10/100 datasets with the overlap parameter $K=\{1,4\}$.
As presented in \tabref{table:ensemble10}, AMCL significantly enhances the oracle and top-1 error rates compared to the previous MCL-based methods.
We found that under the setting of $K=4$, AMCL provides $12.49\%$ and $6.10\%$ relative reductions in the oracle and top-1 error rates from the second best result on the CIFAR-100 dataset.

AMCL seems to take more advantages of larger $M$ compared to the other methods. 
This is because the increase in the ensemble size $M$ causes more chances of misassigned inputs for the previous methods while AMCL does not experience these misassignments due to the memory-based assignment procedure presented in Section 3.2. 

\begin{table*}[ht!]
	\centering
	\begin{tabular}{c c l l | l l }
		\hline
		\multirow{1}{*}{Ensemble} &
		\multirow{2}{*}{$K$} &
		\multicolumn{2}{c}{CIFAR-10} &
		\multicolumn{2}{c}{CIFAR-100} \\ \cline{3-6}
		\multirow{1}{*}{Method} & & Oracle & Top-1 & Oracle & Top-1 \\ \hline
		IE                          & - & 4.20$_{\pm \tx{0.08}}$             & 12.11$_{\pm \tx{0.11}}$ 			& 16.49$_{\pm \tx{0.19}}$ 		& \blue{34.26$_{\pm \tx{0.14}}$} \\ \cline{1-6}
		\multirow{2}{*}{sMCL}		& 1 & \tbf{0.00}$_{\pm \tbf{0.00}}$      & 70.23$_{\pm \tx{0.15}}$ 			& 24.61$_{\pm \tx{0.12}}$ 		& 68.54$_{\pm \tx{0.43}}$ \\
		& 4 & 1.54$_{\pm \tx{0.07}}$			 & 21.89$_{\pm \tx{0.13}}$ 			& 15.78$_{\pm \tx{0.11}}$ 		& 41.69$_{\pm \tx{0.28}}$ \\ \cline{1-6}
		\multirow{2}{*}{CMCL}       & 1 & 2.67$_{\pm \tx{0.11}}$			 & 12.91$_{\pm \tx{0.12}}$ 			& 23.31$_{\pm \tx{0.08}}$ 		& 43.71$_{\pm \tx{0.14}}$ \\
		& 4 & 1.22$_{\pm \tx{0.06}}$			 & \blue{9.24$_{\pm \tx{0.15}}$}	& \blue{14.09$_{\pm \tx{0.17}}$}& 34.67$_{\pm \tx{0.24}}$ \\ \cline{1-6}
		\multirow{2}{*}{vMCL}       & 1 & 2.13$_{\pm \tx{0.10}}$			 & 14.36$_{\pm \tx{0.13}}$ 			& 22.42$_{\pm \tx{0.13}}$ 		& 45.36$_{\pm \tx{0.28}}$ \\
		& 4 & 1.39$_{\pm \tx{0.11}}$			 & 11.62$_{\pm \tx{0.09}}$ 			& 15.14$_{\pm \tx{0.15}}$ 		& 36.13$_{\pm \tx{0.37}}$ \\ \cline{1-6}
		\multirow{2}{*}{AMCL}       & 1 & \blue{0.44$_{\pm \tx{0.09}}$}		 & 10.44$_{\pm \tx{0.11}}$ 			& 19.73$_{\pm \tx{0.11}}$ 		& 41.20$_{\pm \tx{0.13}}$ \\
		& 4 & 0.54$_{\pm \tx{0.07}}$			 & \tbf{8.26}$_{\pm \tbf{0.12}}$~(\tbf{-10.61\%}) & \tbf{12.33}$_{\pm \tbf{0.12}}$~(\tbf{-12.49\%}) & \tbf{32.17}$_{\pm \tbf{0.16}}$~(\tbf{-6.10\%})  \\ \hline
	\end{tabular}
	\caption{The classification error rates~(\%) of ResNet-10 on CIFAR-10/100 datasets with the overlap parameter $K=1,4$ at ensemble size $M=10$. We report the mean and standard deviation~(as subscript) by repeated 3 times. The best scores are marked in \textbf{bold}, and the values in parentheses represent the relative reductions from the second best result in blue.}
	\label{table:ensemble10}
\end{table*}


\vspace{1.5em}
\begin{flushleft}
{\Large \textbf{E. Network Architecture and Additional Results for the Foreground-background Segmentation}}
\end{flushleft}
In this section, the details about the network used for segmentation and the examples of the segmentation results are presented as a supplementary to Section 4.3. 

\begin{figure*}[ht]
	\centering
	\includegraphics[width=0.8\columnwidth, height=0.17\columnwidth]{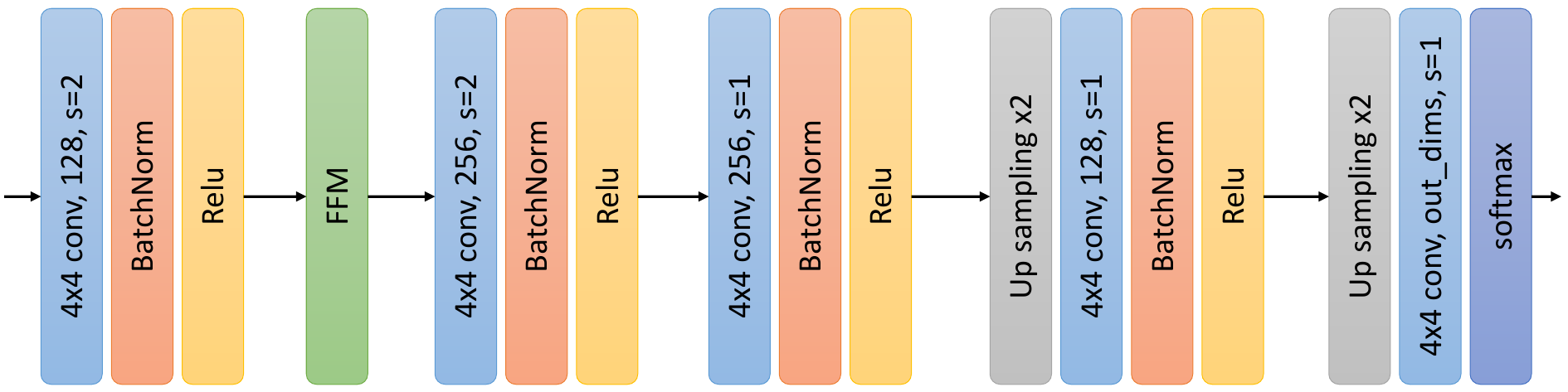}
	\caption{The detailed architecture of the network used for the segmentation task.} 
	\label{fig:segnet}
\end{figure*}

\begin{flushleft}
{\large \textbf{E.1 Network Architecture}} 
\end{flushleft}
\vspace{-0.5em}
The fully convolutional network, used in the foreground-background segmentation task, is illustrated in \figref{fig:segnet}.
The structure of the network consists of encoding and decoding parts.
The encoding part has two convolution-batch normalization-ReLU block operations with 128 and 256 filters consisting of $4 \times 4$ receptive fields and strides of 2. 
Similarly, the decoding part consists of transformed structures of the encoding part with a bi-linear upsampling layer. 
Our feature fusion module (FFM) is inserted after the first encoding block in order to moderate the overconfident issue by collecting features from each model in the lower layers.

\begin{figure*}[ht!]
	\centering
	\includegraphics[width=1.0\columnwidth, height=0.29\columnwidth]{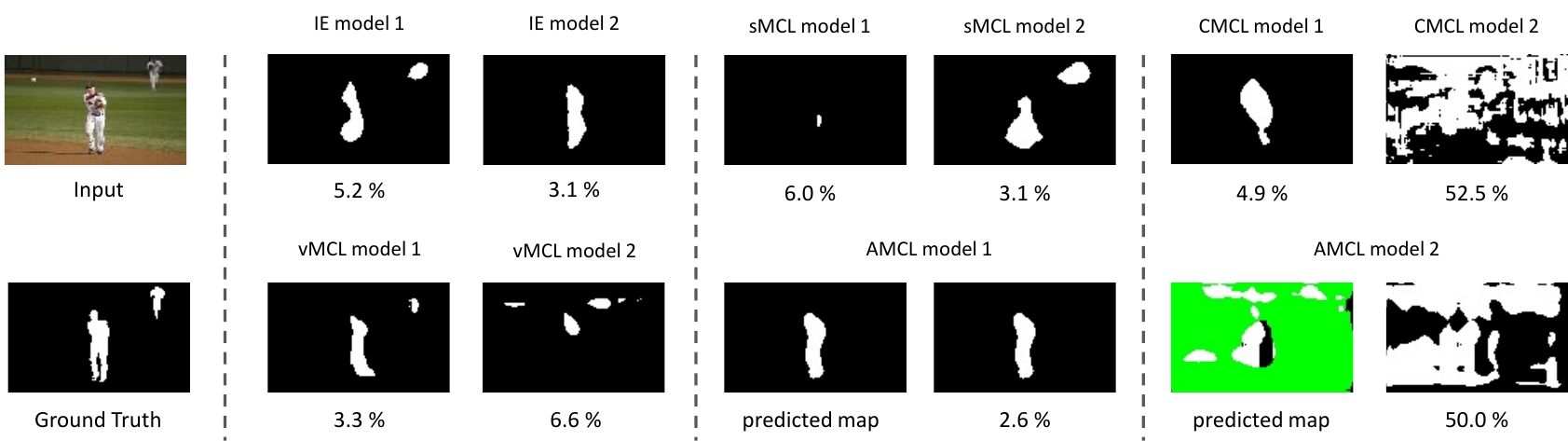} 
	\includegraphics[width=1.0\columnwidth, height=0.29\columnwidth]{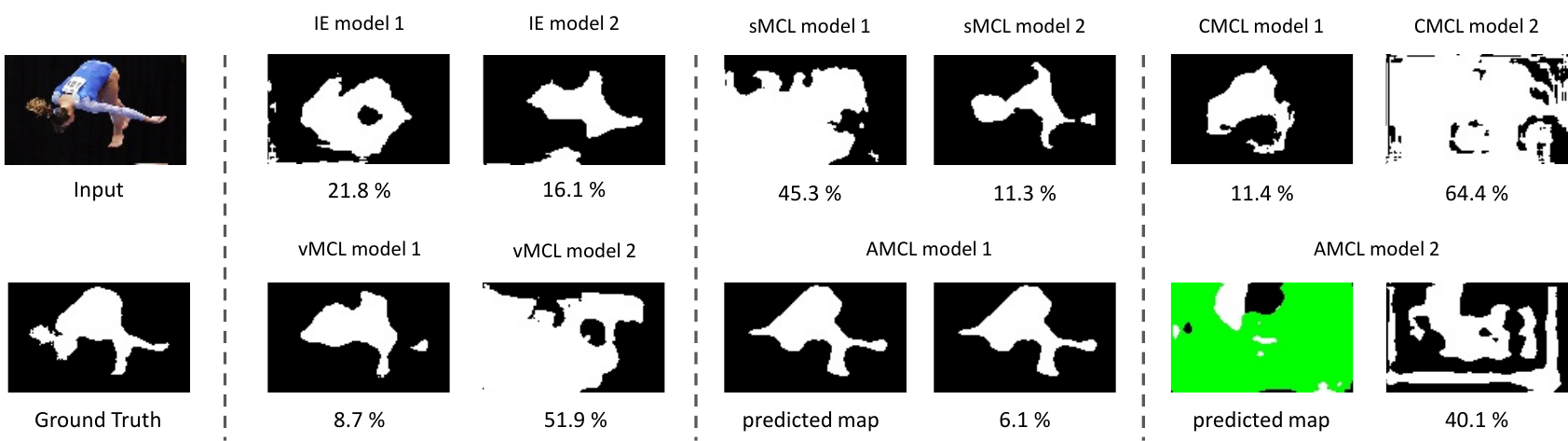} 
	\includegraphics[width=1.0\columnwidth, height=0.29\columnwidth]{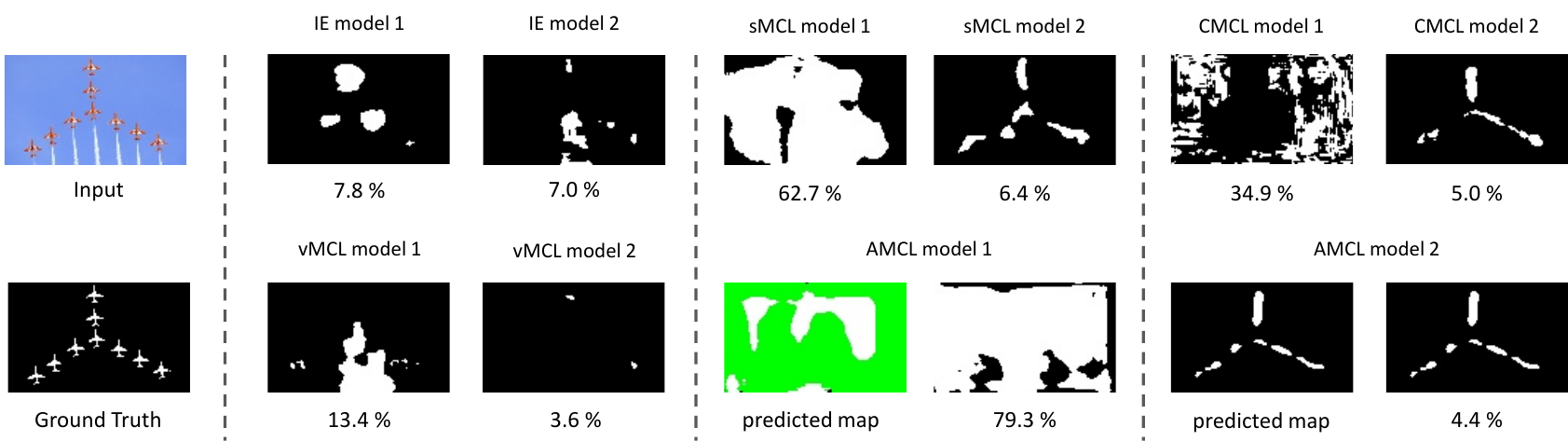} 
	\caption{Three samples of segmentation results with all the variants of MCL, including IE. The three samples are from the classes of Red Sox players, gymnastics, and airshow-plane from top to bottom.} 
	\label{fig:seg_results}
\end{figure*}

\begin{flushleft}
{\large \textbf{E.2 Visualization of the Segmentation Results}} 
\end{flushleft}
\vspace{-0.5em}
The predictions of the foreground-background segmentation with different ensemble methods are presented with selected test images from the iCoseg dataset in \figref{fig:seg_results}. 
In the AMCL results, the pixels covered in green indicate that they are classified as auxiliary class, meaning that the model is not specialized in deciding whether those particular pixels are foreground or background. 
The difference in specialization is apparent between two models of the ensemble with AMCL that results in larger gaps in error rates (2.6$\%$ vs 50.0$\%$ in the sample of Red Sox players, 6.1$\%$ vs 40.1$\%$ in the sample of gymnastics, and 79.3$\%$ vs 4.4$\%$ in the sample of airshow-plane).
The last example with an airshow image is a case in which AMCL is the second best result in terms of the error rate, which is the percentage of incorrectly labeled pixels. 
However, the prediction output looks semantically better than that of vMCL, which is the method with the best result in this example of the airshow image.
Therefore, we can discover that AMCL overall achieves both quantitative and qualitative results that outperform other variant methods of MCL. 

